\PassOptionsToPackage{table}{xcolor}
\documentclass{article} 
\usepackage{iclr2027_conference,times}

\usepackage{hyperref}
\usepackage{xurl}
\usepackage{graphicx}
\usepackage[T1]{fontenc}
\usepackage{booktabs}
\usepackage{amsfonts}
\usepackage{nicefrac}
\usepackage{microtype}
\usepackage[table]{xcolor}
\definecolor{linkskyblue}{HTML}{3AA0D8}
\hypersetup{colorlinks=true,allcolors=linkskyblue}
\usepackage{amsmath}
\usepackage{amssymb}
\usepackage{mathtools}
\usepackage{amsthm}
\usepackage{multirow}
\usepackage{wrapfig}
\usepackage{lipsum}
\usepackage{algorithm}
\usepackage{enumitem}
\usepackage{tabularx}
\usepackage{balance}
\usepackage[noend]{algpseudocode}
\usepackage{pgfplots}
\usepackage{tikz}
\usetikzlibrary{pgfplots.groupplots}
\usetikzlibrary{patterns}
\usepackage{pgf-pie}
\usepackage{adjustbox}
\usepackage{colortbl}
\usepackage{xspace}
\usepackage{pifont}
\usepackage{float}
\usepackage[most]{tcolorbox}
\usepackage{caption}

\definecolor{tableheader}{HTML}{EAF0F4}
\definecolor{tablerow}{HTML}{F7F9FB}
\AtBeginDocument{\raggedbottom}

\renewcommand{\bibfont}{%
  \raggedright
  \urlstyle{same}%
  \Urlmuskip=0mu\relax
}

\newcommand{\benchname}{RefCraft\xspace}

\title{\texorpdfstring{\raisebox{-0.25em}{\includegraphics[height=1.35em]{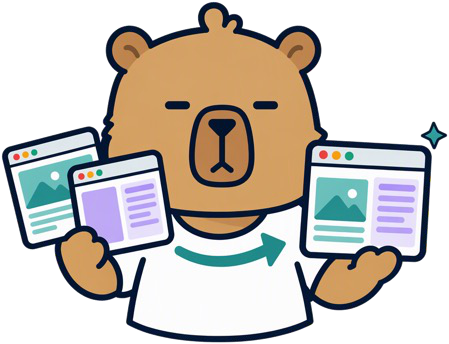}}\hspace{0.25em}}{}RefCraft: Benchmarking Visual-Reference-Conditioned Agentic Generation through Web Development}

\newcommand*\samethanks[1][\value{footnote}]{\footnotemark[#1]}

\author{Yuchi Wang$^{12}$\thanks{~~Work done during an internship at Alibaba ATH.}, ~Rongyao Fang$^{2}$, ~Weihao Yu$^{23}$, ~Xilin Wei$^{24}$, ~Tiancheng Gu$^{25}$, Dingkang Yang$^{14}$, \\ \textbf{Hang Zhang$^{2}$, ~Dayiheng Liu$^{2}$, ~Hongsheng Li$^{1}$\thanks{~~Hongsheng Li and Shuai Bai are corresponding authors.}, ~Shuai Bai$^{2}$\samethanks} \\
 {$^{1}$MMLab, CUHK $~~~$ $^{2}$Alibaba Token Foundry $~~~$ $^{3}$The Chinese Univerity of Hong Kong}
 \\
    {$^{4}$Fudan University $~~~$  
    $^{5}$The University of Sydney
    }\\
     \texttt{wangyuchi369@gmail.com~hsli@ee.cuhk.edu.hk}\\
     \texttt{baishuai.bs@alibaba-inc.com} \\
   ~\\
   \makebox[\dimexpr\textwidth-2\tabcolsep\relax][c]{\normalfont\normalsize\bfseries\url{https://wangyuchi369.github.io/RefCraft/}}\\
}

\iclrfinalcopy 
\iclrarxivcopy 
\begin{document}

\maketitle

\begin{figure*}[h]\centering
\includegraphics[width=1\textwidth]{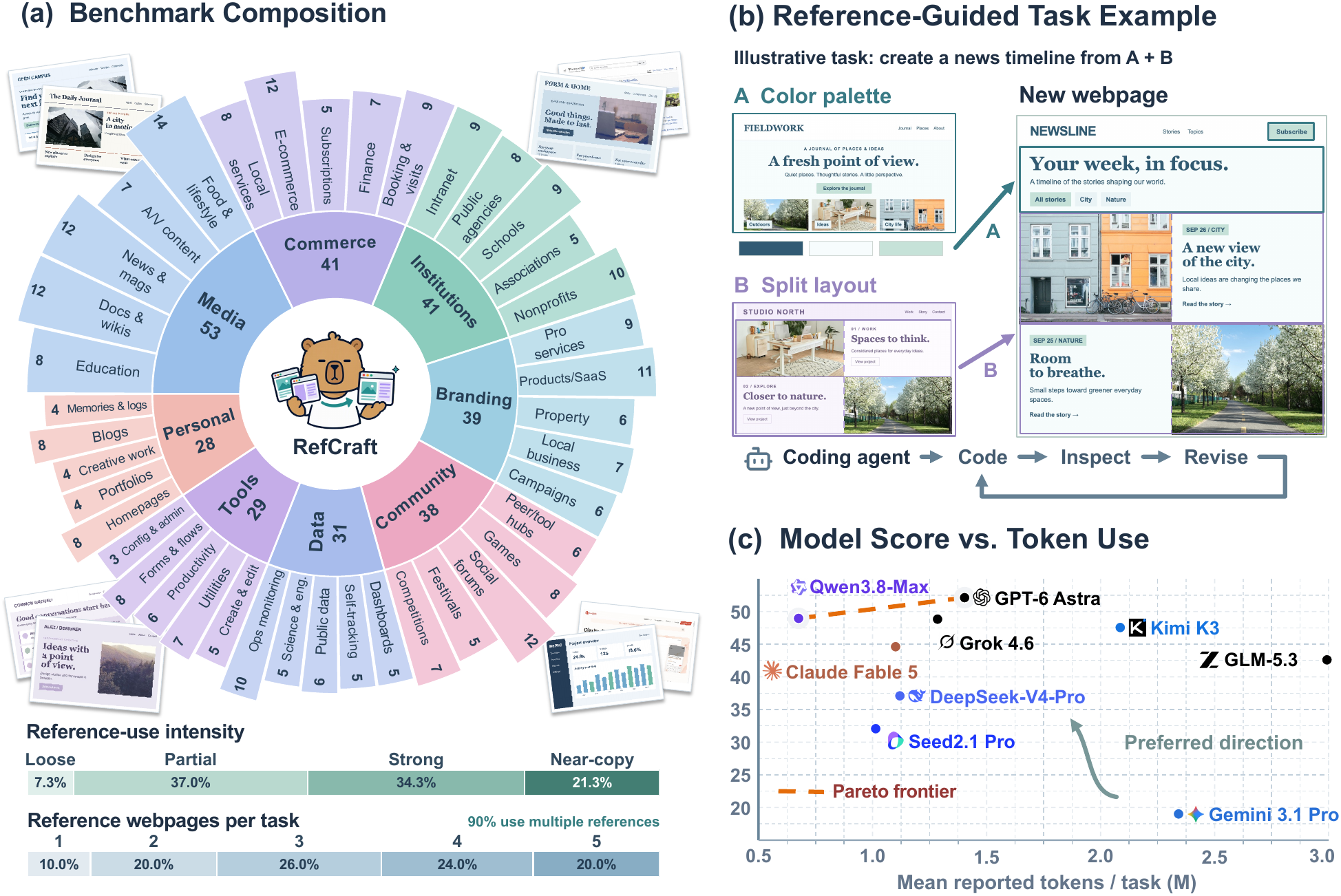}
\vspace{-7pt}
\caption{\textbf{Overview of \benchname.} \textbf{(a)} \benchname spans a broad spectrum of practical web development domains and features visual references that vary in intensity and source. (b) An example task in which a new webpage borrows the color palette from source webpage A and the layout from source webpage B. The agent may iteratively render, inspect, and revise the webpage. (c) Overall score versus reported token usage for nine frontier models. The dashed line denotes the Pareto frontier. The results show that the task remains far from solved.}
\label{fig:teaser}
\end{figure*}

\begin{abstract}
Beyond basic tool use and coding skills, effectively incorporating visual references into agentic generation is equally essential for applying AI agents to real-world productivity tasks. We study this crucial capability through webpage development, a practical and readily evaluable setting in which existing benchmarks remain largely idealized: they typically focus on reconstructing visual targets or generating webpages from textual instructions alone, overlooking the more common workflow of developing from visual examples. Specifically, we introduce \benchname, a benchmark comprising 300 realistic multimodal webpage-development tasks that combine natural-language instructions with screenshots from one to five reference webpages. Agents must selectively interpret and utilize these references while rendering, visually inspecting, and iteratively refining their outputs. We evaluate the resulting open-ended artifacts using independently verifiable keypoints and complementary programmatic and VLM judges. Across nine frontier (M)LLMs, the strongest system achieves an overall score of only 52.12 out of 100, revealing that visual-reference-conditioned agentic generation remains a substantial challenge for current models in practical webpage-development settings.
\end{abstract}

\section{Introduction}
\label{sec:intro}

\begingroup
\small
\setlength{\parskip}{0pt}
\raggedleft
\textit{``Creativity is just connecting things.''}\\*[-1pt]
--- Steve Jobs\par
\endgroup
\vspace{-2pt}

Recently, as coding and co-working assistants, (M)LLMs have transformed numerous fields and substantially improved productivity~\citep{openai2026gpt55,anthropic2026fable5,qwen38,team2026kimi}. This success has further fueled growing enthusiasm among researchers and practitioners for extending large models to an even broader range of practical tasks, as exemplified by GPT-6 Astra~\citep{openai2026gpt6astra} and Claude Fable 5.1~\citep{anthropic2026fable51}. When we scrutinize some common productivity tasks, we find that \textbf{visual references} play an important role: designers may first study successful examples before developing new proposals and then continuously refine their designs through visual inspection, while CAD engineers examine existing parts to design new ones that fit and function properly. This observation highlights that, for both human experts and AI agents, \emph{basic tool use and coding skills alone do not encompass the full spectrum of productive capabilities---effectively \textbf{incorporating visual reference and inspection} into agentic generation is equally essential}.

To evaluate this important yet subtle capability, we ground it in a focused and readily verifiable testbed: webpage development, which is ubiquitous in everyday digital life and produces highly visual, easily assessable outputs. In fact, a growing body of benchmarks has investigated general webpage-related tasks, but most either use natural-language instructions without visual references~\citep{lu2026webgen,zhu2025frontendbench} or focus solely on reconstruction tasks, such as reconstructing webpages from screenshots~\citep{yun2024web2code,wang2026webgen} or videos~\citep{chen2026iwr}. More recent studies have broadened the evaluation setting by incorporating interactive operations~\citep{xiao2025interaction2code} or introducing webpage editing and repair tasks~\citep{lei2026webcompass}. Nevertheless, we argue that \textbf{these prior efforts remain academically idealized and differ considerably from common real-world development workflows}. In practice, frontend developers and users often begin with a rough design idea and browse existing websites for inspiration. For instance, a user may adopt the appealing layout of a reference website while redesigning its fonts, content, and other elements for a new purpose. Fully reconstructing an existing website is a far less common or representative task. To bridge this gap and, more importantly, evaluate whether AI agents can effectively conduct such a visual-reference-conditioned agentic generation process, we introduce \benchname.

\benchname comprises 300 diverse and realistic multimodal queries spanning a broad range of web development scenarios. Each query combines a natural-language user instruction with screenshots from one to five reference webpages, and models solve the task in an inspect-and-iterate setting where they can render and refine intermediate outputs. As shown in Fig.~\ref{fig:teaser}, \benchname varies substantially in domains, reference composition, and the intended degree of reference utilization---from closely following a reference to selectively adapting its overall style or specific local elements. Evaluating such tasks requires moving beyond conventional target-based evaluation: a query may admit multiple valid solutions, and visual similarity alone cannot determine whether the references have been used appropriately. We therefore decompose each complex and subtle query into a verified set of key points, turning open-ended requirements into independently assessable criteria. Moreover, webpages are both visual and interactive artifacts, encompassing subjective perceptual qualities as well as objective executable behaviors. We thus adopt a dual-track evaluation protocol that uses programmatic judges for deterministic requirements and a VLM judge for perceptual and semantic ones. Together, these designs enable reliable evaluation of whether a generated webpage fulfills the user's intended use of visual references, rather than merely resembling them.

We benchmark 9 frontier (M)LLMs on \benchname under a unified harness. Despite access to visual references and an inspect-and-iterate workflow, the best-performing system, GPT-6 Astra~\citep{openai2026gpt6astra}, achieves an overall score of only 52.12, leaving substantial room for improvement. Component-level results reveal that models often satisfy many individual requirements but fail to realize an essential reference relationship or core behavior. To assess the reliability of \benchname, we compare its model ranking with both our expert human study and the human-preference leaderboard of Arena.AI\footnote{\url{https://arena.ai/leaderboard/code/image-to-webdev}}, finding strong agreement with both. These findings indicate that current agents can reproduce salient visual attributes but still struggle to translate instruction-dependent reference relationships into coherent designs---precisely the capability that matters to human users.


\section{Related Work}

\subsection{Visual-Reference-Conditioned Agentic Generation}

%
%

In recent years, we have witnessed rapid advances in AI agents. From early frameworks such as ReAct~\citep{yao2022react} to coding agents such as SWE-agent~\citep{yang2024swe} and OpenHands~\citep{wang2025openhands}, substantial effort has been devoted to reasoning, code generation, long-horizon planning, and tool use~\citep{schick2023toolformer,wang2024executable,Yang2026WhenLP}. In comparison, vision has received less attention in agentic generation and is typically used only as a perceptual interface for interpreting image or video observations~\citep{xie2024osworld,qin2025ui,chen2024pca}.
Meanwhile, visual inputs serve as important conditioning signals in generative models, particularly diffusion-based models for image and video generation~\citep{seedance2026seedance,bai2025uniedit,wang2025instructavatar}. Yet methods ranging from DreamBooth~\citep{ruiz2023dreambooth} and ControlNet~\citep{zhang2023adding} to IP-Adapter~\citep{ye2023ip} and the OmniGen series~\citep{xiao2025omnigen,wu2025omnigen2} often assign predefined, coarse-grained roles to visual references---for example, preserving subject identity, transferring overall style, or imposing structural constraints. Consequently, both prominent research directions leave a challenging yet practical setting underexplored: visual-reference-conditioned agentic generation, where an agent must interpret subtle user instructions and selectively incorporate heterogeneous visual references throughout the agentic generation process.

\subsection{Web Development Benchmarks}

%
%
%
%
%
%
%
%

Existing web development benchmarks broadly follow two task formulations. Text-driven benchmarks, such as WebGen-Bench~\citep{lu2026webgen} and FrontendBench~\citep{zhu2025frontendbench}, ask models to construct webpages or applications from natural-language specifications~\citep{tran2026vibe,xu2025web}. Visual-target benchmarks instead formulate generation as reconstruction, including screenshot-to-code tasks such as Design2Code~\citep{si2025design2code} and Web2Code~\citep{yun2024web2code}, video- or interaction-conditioned generation~\citep{chen2026iwr,xiao2025interaction2code}, and sketch-to-code generation~\citep{jain2019sketch2code}. More recent benchmarks, including DesignBench~\citep{xiao2025designbench}, ArtifactsBench~\citep{zhang2025artifactsbench}, and WebCompass~\citep{lei2026webcompass}, extend beyond one-shot generation to interactive behavior, editing, repair, and multi-turn refinement~\citep{yang2025ui2code,sun2025fullfront,saxena2026swe}.
Evaluation for these tasks has evolved accordingly, moving from visual-similarity measures based on CLIP features~\citep{radford2021learning}, layout blocks, colors, and positions~\citep{wang2026webgen,si2025design2code} toward MLLM- or agent-as-a-judge evaluation over rendered webpages~\citep{zhang2025artifactsbench,lei2026webcompass,zhuge2024agent}. Nevertheless, existing benchmarks typically treat visual input as a target to reconstruct or an artifact to edit. They rarely capture the common workflow in which users consult one or more webpages as partial design references, which is precisely the setting studied in our work.

\begin{figure*}[h]\centering
\includegraphics[width=0.98\textwidth]{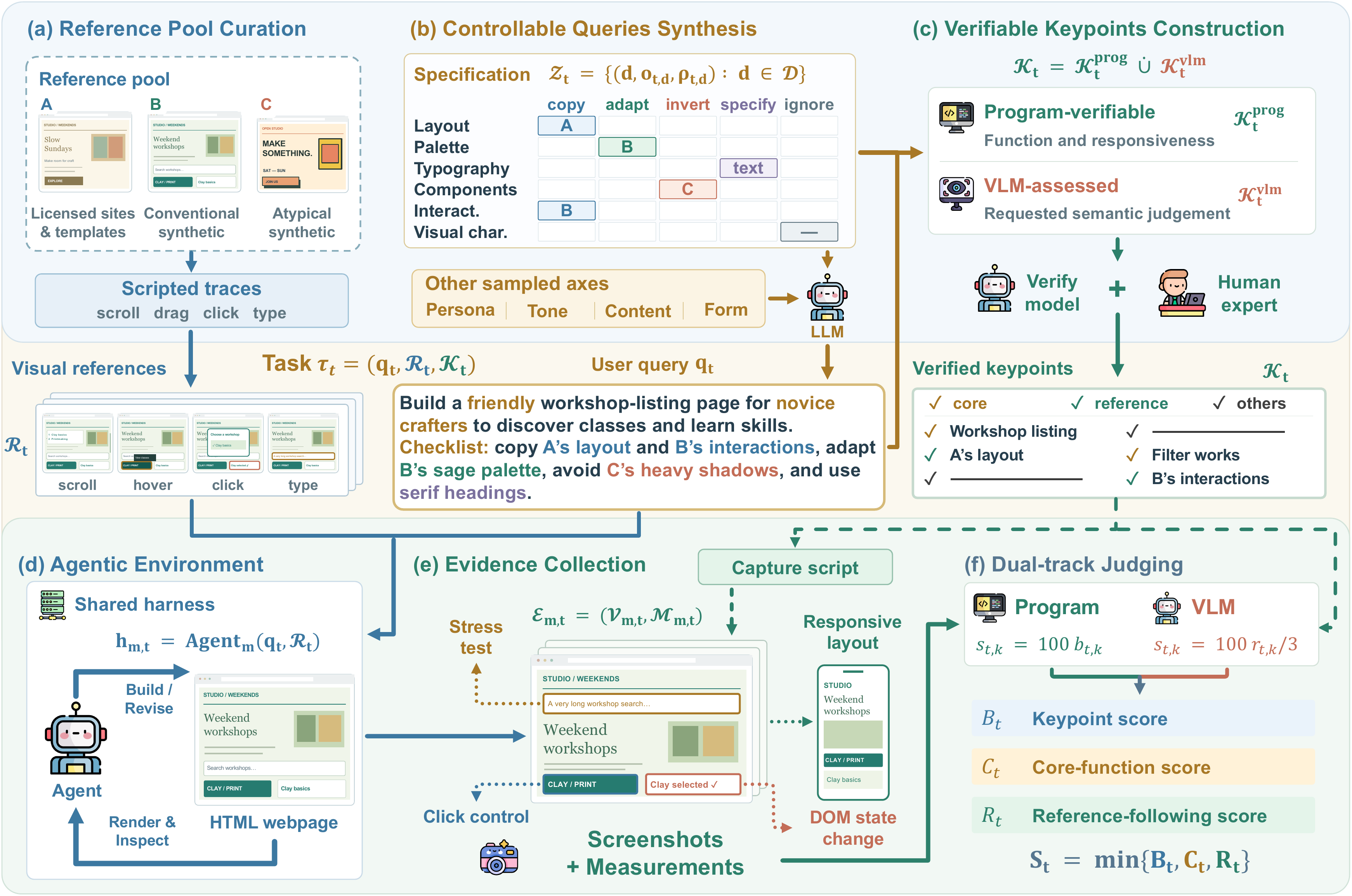}

\caption{\textbf{Construction and evaluation of \benchname.} (a) Reference webpages are captured across interaction states. (b) A specification spanning six visual dimensions and five requirement types guides natural-language query synthesis. (c) Task requirements are decomposed into program-verifiable and VLM-assessed keypoints. (d) Given the query and reference images, agents iteratively build, render, inspect, and revise a webpage in a shared harness. (e) Keypoints guide evidence collection from generated artifacts. (f) Complementary programmatic and VLM judges produce the base keypoint score $B$, core-function score $C$, and reference-following score $R$.}
\label{fig:bmk}
\end{figure*}


\section{Benchmark Design}
\label{sec:benchmark}

Fig.~\ref{fig:bmk} presents the overall design of \benchname. Each benchmark instance, corresponding to a single task, consists of a textual user query, one to five visual references, and a verified set of evaluation keypoints. We construct these instances in three stages: curating a diverse reference pool (\textsection~\ref{sec:reference-pool}), synthesizing controllable user queries (\textsection~\ref{sec:query-synthesis}), and decomposing each query into independently verifiable keypoints (\textsection~\ref{sec:keypoint-construction}). Further implementation details  are provided in Appendix~\ref{app:implementation-details}.



\subsection{Task Formulation}
\label{sec:task-formulation}

Each task $t$ in \benchname consists of a natural-language instruction $q_t$ and a set of visual references $\mathcal{R}_t = \{R_i\}_{i=1}^{N_t}$, where $1 \leq N_t \leq 5$ and each $R_i$ contains screenshots sampled from a reference webpage. Together with the evaluation keypoints $\mathcal{K}_t$, they define the task $\tau_t=(q_t,\mathcal{R}_t,\mathcal{K}_t)$. Model $m$ executes the task and produces $h_{m,t}=\operatorname{Agent}_m(q_t,\mathcal{R}_t)$, a self-contained HTML webpage, without access to $\mathcal{K}_t$. During generation, the agent may render its intermediate implementation, visually inspect the result, and iteratively revise the webpage. Unlike screenshot reconstruction, the objective is not to reproduce any reference exactly. Depending on the instruction, the agent may need to closely follow one reference, combine properties from multiple references, adapt selected design elements, or deliberately avoid particular characteristics. The task therefore evaluates whether an agent can interpret and appropriately utilize visual references within an inspect-and-iterate workflow.

\subsection{Reference Pool Curation}
\label{sec:reference-pool}

Reference webpages are important because they determine what information an agent can draw upon and therefore form the foundation of \benchname. We curate the reference webpage pool according to three principles: \textbf{diversity}, covering a broad range of topics and design styles; \textbf{visual richness}, providing sufficiently salient properties for users to reference and evaluators to verify; and \textbf{realism}, reflecting designs that may plausibly appear in practical development workflows.
Following these principles, we combine three sources: \textbf{(1)} openly licensed websites and webpage templates; \textbf{(2)} conventionally styled webpages synthesized by multiple models to increase diversity; and \textbf{(3)} deliberately synthesized webpages that deviate from conventional modern styles and common model-generated patterns. The rationale is that synthesized pages may inherit common model priors, allowing an agent to produce superficially similar designs based on its prior knowledge without consulting the references. To mitigate this shortcut, the third group incorporates less conventional visual characteristics, such as unusual color palettes, uncommon font combinations, and hard-edged shadows.

We represent each reference using images rather than source code or a live URL, preventing implementation details from leaking to the evaluated agent and better reflecting practical scenarios.  Because a single static screenshot cannot expose dynamic properties such as hover effects or interactive state changes, we use Playwright\footnote{\url{https://playwright.dev/}} to capture a sequence of visual frames along scripted interaction traces involving scrolling, hovering, clicking, typing, and key presses.
To support subsequent task construction, we also profile the observable information in each reference along six dimensions: \emph{layout}, \emph{palette}, \emph{typography}, \emph{components}, \emph{interaction}, and \emph{overall visual character}. Each dimension is annotated as \emph{rich}, \emph{weak}, or \emph{none}, together with a concise description of the observed evidence. This profile prevents later queries from referring to properties that are absent or visually ambiguous.

\subsection{Controllable Query Synthesis}
\label{sec:query-synthesis}

Directly prompting an LLM to generate user queries leads to substantial mode collapse: the resulting instructions are often similarly phrased and repeatedly emphasize only a small subset of the available reference information. We instead adopt a \textbf{specification-first} approach: we first determine the intended reference usage and task composition and then verbalize them as a natural user request.

\paragraph{Reference-use specification.}
As shown in Fig.~\ref{fig:bmk}(b), let $\mathcal{D}$ denote the six visual dimensions above. We represent the reference-use specification for task $t$ as $\mathcal{Z}_t=\{(d,o_{t,d},\rho_{t,d}):d\in\mathcal{D}\}$. For each dimension, $o_{t,d}\in\{\emph{copy},\emph{adapt},\emph{invert},\emph{specify},\emph{ignore}\}$ specifies the requirement type, while $\rho_{t,d}$ identifies the assigned reference, with $\rho_{t,d}=\varnothing$ for \emph{specify} and \emph{ignore}. The five types are \emph{copy}, which closely follows the reference; \emph{adapt}, which preserves only selected characteristics; \emph{invert}, which deliberately avoids the referenced characteristic; \emph{specify}, which follows an explicit textual requirement rather than a reference; and \emph{ignore}, which leaves the dimension unconstrained. We retain a sampled reference-dependent requirement only when the assigned reference contains sufficiently clear and discriminable evidence for the corresponding dimension. 

\paragraph{Compositional query design.}
We construct a design matrix whose axes describe the user persona, target content, form and tone of the request, selected references, the reference-use specification $\mathcal{Z}_t$, and other relevant factors. Values are sampled across these axes and subsequently filtered to remove contradictory, implausible, or otherwise pathological combinations. An LLM then converts each valid combination into a fluent natural-language instruction. This procedure preserves linguistic naturalness while maintaining systematic control over task diversity. Additional generation details and examples are provided in Appendix~\ref{app:query-generation}.

\subsection{Verifiable Keypoint Construction}
\label{sec:keypoint-construction}

A query in \benchname may admit many valid implementations, and its requirements can be fine-grained, subtle, and distributed across both language and visual references. Consequently, evaluating against a single canonical webpage or a holistic similarity score would be inappropriate. Inspired by rubric-based evaluation~\citep{yu2026specv, hashemi2024llm, pathak2025rubric}, we instead decompose each query into a set of non-overlapping keypoints, each capturing one independently assessable requirement.
We partition $\mathcal{K}_t=\mathcal{K}_t^{\mathrm{prog}}\mathbin{\dot\cup}\mathcal{K}_t^{\mathrm{vlm}}$ into disjoint tracks according to how each keypoint can be most reliably verified. \emph{Program-verifiable keypoints $\mathcal{K}_t^{\mathrm{prog}}$}  capture deterministic properties such as color distance, contrast, responsive behavior, DOM state changes, and robustness to abnormal content. The remaining keypoints require semantic or perceptual judgment and are assigned to the \emph{VLM-assessed track $\mathcal{K}_t^{\mathrm{vlm}}$}. We follow the principle that whenever reliable deterministic verification is possible, we favor the programmatic track to reduce reliance on VLM judgments, which can be overly lenient. We additionally designate essential user requirements and page-integrity constraints as critical items, ensuring that they receive greater emphasis in the final evaluation.

We apply a multi-stage validation procedure to ensure keypoint quality. A separate model first reviews the full set of keypoints for completeness and may add missing requirements. It then examines each keypoint independently to identify hallucinated, redundant, or unverifiable criteria. Finally, ten web development experts review and revise the resulting keypoints. This process produces a verified evaluation specification for every benchmark instance.


\section{Evaluation Protocol}
\label{sec:protocol}

Given the benchmark instances constructed above, our evaluation protocol standardizes how models interact with the task, converts their unconstrained HTML outputs into structured evidence, and scores them against the corresponding keypoints.

\subsection{Unified Agentic Environment}
\label{sec:evaluation-environment}

We evaluate each model as a coding agent, reflecting the increasingly common use of agentic harnesses in practical development workflows~\cite{}. To ensure comparability, all models operate through the same open-source coding harness, \texttt{opencode}~\footnote{\url{https://github.com/anomalyco/opencode}}, with identical tools, workspace structure, and output requirements.
For each task, the agent receives the user instruction, visual references, and an environment brief. It is given a dedicated working directory and may render, screenshot, and inspect intermediate implementations before revising them. The final result must be stored as a single self-contained HTML file, ensuring a uniform artifact format across models.
We restrict access that could reveal benchmark internals or ground-truth evaluation criteria. Each working directory is isolated from the benchmark repository, and attempts to access the repository or evaluation keypoints are denied. Network access remains available to preserve a realistic coding environment, but searching for or downloading the original source of a reference webpage invalidates the run. 

\subsection{Evidence Collection}
\label{sec:evidence-collection}

Generated HTML files vary substantially in structure and may exhibit visual and interactive behaviors that cannot be reliably evaluated from source code alone. Therefore, for each artifact $h_{m,t}$, we collect structured evidence $\mathcal{E}_{m,t}=(\mathcal{V}_{m,t},\mathcal{M}_{m,t})$, where $\mathcal{V}_{m,t}$ contains captured views and $\mathcal{M}_{m,t}$ contains browser measurements. Specifically, for each benchmark instance, we derive a task-specific capture script specifying a sequence of actions, such as scrolling, hovering, clicking, and typing. The script is designed to expose the page states required to evaluate as many relevant keypoints as possible. We load the generated HTML in Playwright, execute the actions sequentially, and capture screenshots of the resulting states.
In parallel, we apply a fixed suite of browser-based measurements to assess properties such as rendering integrity, responsive behavior, robustness to abnormal content, and whether interactive controls produce genuine state changes. Together, the screenshots and measurements provide downstream judges with targeted evidence rather than requiring them to infer page behavior from raw HTML, which would be poorly aligned with practical usage.

\subsection{Dual-Track Judging and Score Aggregation}
\label{sec:scoring}

Our evaluation first produces keypoint-level scores from $\mathcal{E}_{m,t}$ using complementary programmatic and VLM judges. The programmatic judge compares browser measurements against deterministic keypoints and returns a binary verdict. For keypoints requiring semantic or perceptual judgment, the VLM judge receives the keypoint, relevant screenshots, and supporting measurements, and assigns a rating on a four-level scale from $0$ to $3$. Each keypoint is evaluated independently to reduce interference between unrelated requirements. Denote these judgments by $b_{t,k}\in\{0,1\}$ and $r_{t,k}\in\{0,1,2,3\}$, respectively, omitting the model index for clarity. Both tracks are normalized to the same percentage scale: $s_{t,k}=100\,b_{t,k}$ for $k\in\mathcal{K}_t^{\mathrm{prog}}$ and $s_{t,k}=100\,r_{t,k}/3$ for $k\in\mathcal{K}_t^{\mathrm{vlm}}$.

We aggregate these keypoint-level judgments into three complementary task-level quantities. The \textbf{base keypoint score $B_t$} is the mean of all normalized keypoint scores:
$B_t=|\mathcal{K}_t|^{-1}\sum_{k\in\mathcal{K}_t}s_{t,k}$.
The \textbf{core-function score $C_t$} captures the weakest essential functionality of the generated webpage. The core goals include renderability and task-specific essential interactions, and $C_t$ is defined as the minimum completion rate across these goals. This metric reflects the intuition that even strong performance on peripheral requirements cannot compensate for a broken primary function, as such failures fundamentally undermine the user experience. 
The \textbf{reference-following score $R_t$} measures whether the generated webpage realizes the requested relationship to its visual references, which is the central capability evaluated by \benchname. We compute $R_t$ by first averaging the keypoint scores within each reference group and then averaging across all applicable groups. 
Because core functionality and reference adherence are both essential to user-perceived quality, $C_t$ and $R_t$ act as gates on the aggregate keypoint score. The final task score therefore takes the minimum over the applicable quantities rather than naively averaging them:

\begin{equation*}
S_t=
\begin{cases}
\min\{B_t,C_t,R_t\}, & \text{if } R_t \text{ is defined},\\
\min\{B_t,C_t\}, & \text{otherwise},
\end{cases}
\qquad
\bar{S}_m=
\frac{1}{|\mathcal{T}_m|}
\sum_{t\in\mathcal{T}_m}S_t,
\end{equation*}

where $\mathcal{T}_m$ denotes the set of tasks evaluated for model $m$. Additional details on judging and score aggregation are provided in Appendix~\ref{app:evaluation-details}.

\section{Experiments}
\label{sec:experiments}

\subsection{Experimental Setup}
\label{sec:experimental-setup}

We evaluate nine frontier models---GPT-6 Astra~\citep{openai2026gpt6astra}, Claude Fable 5~\citep{anthropic2026fable5}, Qwen3.8-Max~\citep{qwen38}, Grok 4.6~\citep{xai2026grok46}, Kimi K3~\citep{team2026kimi}, GLM-5.3~\citep{glm53}, DeepSeek-V4-Pro~\citep{xu2026deepseek}, Seed2.1 Pro~\citep{bytedance2026seed21}, and Gemini 3.1 Pro~\citep{google2026gemini31pro}---as coding agents on \benchname. All models are evaluated using the same \texttt{OpenCode} v1.18.25 harness, with high reasoning effort and a maximum of 100 turns. Keypoints are generated by Claude Opus 5~\citep{anthropicopus5} and verified by GPT-5.6 Sol~\citep{openai2026gpt56sol}. We use Gemini 3.1 Pro as the VLM judge. We report the overall score $S$ together with its three constituent metrics. Further details are provided in Appendix~\ref{app:implementation-details}.

\subsection{Main Results}
\label{sec:main-results}

\begin{table*}[t]
\centering
\small
\setlength{\tabcolsep}{2.5pt}
\renewcommand{\arraystretch}{1.15}
\newcommand{\mainmodellogo}[1]{\raisebox{-0.25ex}{\includegraphics[width=1.8ex,height=1.8ex,keepaspectratio]{figures/logos/logo_#1.pdf}}\hspace{0.4em}}
\caption{\textbf{Results on the 300-task \benchname benchmark.} We report Overall and its three component scores: keypoint completion (Keypts., $B$), core functionality (Core, $C$), and reference following (Ref., $R$). Rollout statistics report mean agent turns, total tokens (Total Tok.), and tool calls per task. Bold and underlined values indicate the best and second-best entries in each column, respectively (higher scores and lower resource counts).}
\label{tab:main-results}
\begin{adjustbox}{max width=\textwidth}
\rowcolors{3}{tablerow}{white}
\begin{tabular}{l*{4}{>{\centering\arraybackslash}m{5.8em}}>{\centering\arraybackslash}m{3.2em}>{\centering\arraybackslash}m{5.6em}>{\centering\arraybackslash}m{3.2em}}
\toprule
& \multicolumn{4}{c}{\textbf{Evaluation}} & \multicolumn{3}{c}{\textbf{Rollout Statistics}} \\
\cmidrule(lr){2-5}\cmidrule(lr){6-8}
\rowcolor{tableheader}
\textbf{Model} & \mbox{\textbf{Overall}$\uparrow$} & \mbox{\textbf{Keypts.}\,$B\uparrow$} & \mbox{\textbf{Core}\,$C\uparrow$} & \mbox{\textbf{Ref.}\,$R\uparrow$} & \textbf{Turns} & \mbox{\textbf{Total Tok.}\textsuperscript{\textdagger}} & \textbf{Calls} \\
\midrule
\mainmodellogo{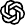}GPT-6 Astra            & \textbf{52.12} & \textbf{75.68} & \textbf{94.89} & 57.93 & 21.7 & 1.40\,M & 30.1 \\
\mainmodellogo{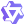}Qwen3.8-Max            & \underline{48.95} & \underline{73.86} & \underline{85.56} & 62.61 & \textbf{9.8} & \textbf{0.68\,M} & \textbf{11.3} \\
\mainmodellogo{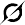}Grok 4.6               & 48.83 & 72.96 & 81.17 & \textbf{66.25} & 21.2 & 1.28\,M & 41.2 \\
\mainmodellogo{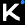}Kimi K3                & 47.55 & 72.74 & 85.00 & 61.33 & 25.4 & 2.09\,M & 31.3 \\
\mainmodellogo{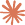}Claude Fable 5         & 44.61 & 72.05 & 77.83 & 61.44 & \underline{15.2} & 1.10\,M & \underline{16.9} \\
\mainmodellogo{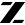}GLM-5.3                & 42.60 & 71.23 & 83.56 & 55.77 & 38.3 & 2.99\,M & 42.2 \\
\mainmodellogo{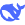}DeepSeek-V4-Pro        & 37.09 & 71.16 & 71.72 & 53.46 & 25.3 & 1.12\,M & 25.6 \\
\mainmodellogo{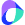}Seed2.1 Pro    & 32.06 & 69.00 & 56.56 & 58.67 & 22.7 & \underline{1.01\,M} & 21.8 \\
\mainmodellogo{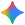}Gemini 3.1 Pro & 19.01 & 61.45 & 33.44 & \underline{65.00} & 26.7 & 2.34\,M & 25.9 \\
\bottomrule
\end{tabular}%
\end{adjustbox}
\par\vspace{3pt}
\begin{minipage}{\textwidth}
\scriptsize\raggedright
\textsuperscript{\textdagger} Token counts are taken from API usage reports and sum input, cache-read, cache-write, output, and reasoning tokens. M denotes millions.
\end{minipage}
\end{table*}

Tab.~\ref{tab:main-results} presents the overall and component scores of the evaluated frontier models, together with efficiency statistics such as token usage and the number of interaction turns. From these results, we could make the following observations:

\paragraph{Visual-reference-conditioned generation remains far from solved.}
As shown in Tab.~\ref{tab:main-results}, GPT-6 Astra achieves the highest overall score of 52.12 out of 100, most models cluster around 40, and Gemini 3.1 Pro scores below 20. These results leave substantial room for improvement. Moreover, the relatively low overall scores contrast with the higher and more concentrated keypoint scores of 61.45--75.68, indicating that models can satisfy many individual requirements yet struggle to produce artifacts that simultaneously preserve essential functionality and realize the user-specified relationship to the references. This gap captures the central challenge posed by \benchname:  a visually plausible webpage cannot be considered successful if it uses references inconsistently with the user's instructions, reflecting how users assess artifacts in real-world settings.

\paragraph{Reference use and functional execution constitute distinct bottlenecks.}
For seven of the nine models, the reference-following score $R$ is the lowest component, identifying reference utilization as the dominant bottleneck. However, reference use alone does not determine end-to-end performance: Gemini 3.1 Pro achieves an $R$ of 65.00 but ranks ninth overall because its core-function score is only 33.44, whereas GPT-6 Astra leads with the strongest core-function score of 94.89 despite a lower $R$. These contrasting failure regimes demonstrate that reliable implementation and appropriate reference utilization capture complementary capabilities, motivating their joint consideration in \benchname.

\paragraph{More budget does not necessarily yield more effective performance.}
Performance is not monotonic in the number of turns, total reported tokens, or tool calls. Qwen3.8-Max outperforms Grok 4.6 while using substantially fewer total tokens (0.68M versus 1.28M), turns, and tool calls. GLM-5.3 uses the most turns, total tokens (2.99M), and tool calls, yet obtains a substantially lower overall score than the leading models. We further present a cost--efficiency Pareto curve in Fig.~\ref{fig:teaser}, which highlights the distinct efficiency profiles of different models.

\begin{wrapfigure}{l}{0.60\textwidth}
\centering
\includegraphics[width=\linewidth]{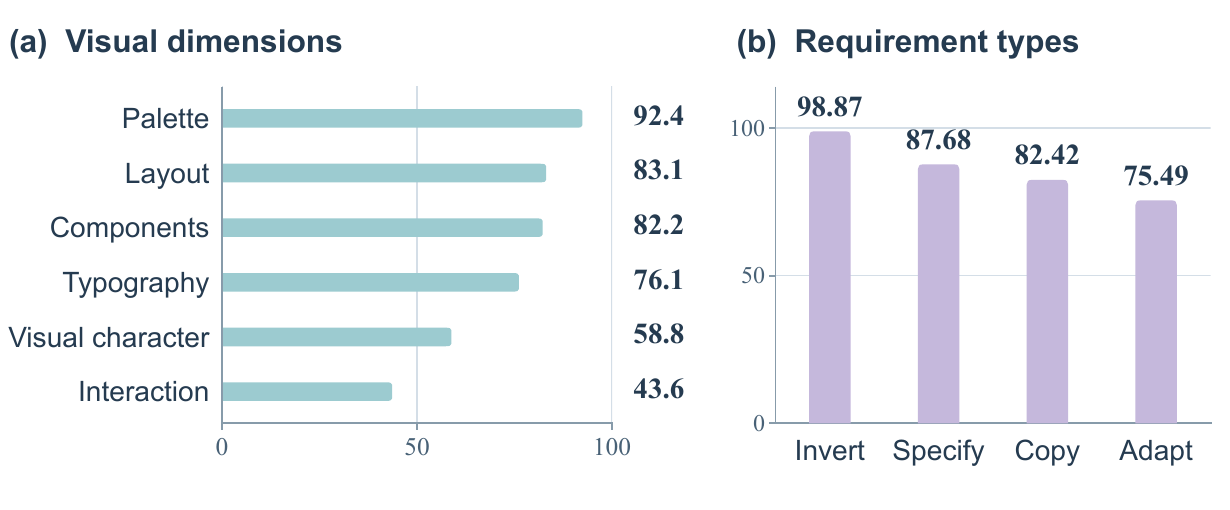}
\caption{Fine-grained performance by (a) visual dimension and (b) requirement type, averaged equally across models.}
\label{fig:fine-grained-analysis}
\vspace{-10pt}
\end{wrapfigure}


\subsection{Further Analysis}

\subsubsection{Fine-Grained Capability Analysis}
\label{sec:fine-grained-analysis}

\begin{table}[t]
\centering
\captionsetup{justification=raggedright,singlelinecheck=false,textfont=normalfont}
\fontsize{7.5}{8.6}\selectfont
\setlength{\tabcolsep}{2.8pt}
\renewcommand{\arraystretch}{1.08}
\colorlet{analysisblue}{black}
\colorlet{analysisband}{tableheader}
\colorlet{analysishighlight}{tablerow}
\begin{minipage}[t]{0.44\linewidth}
\vspace{0pt}
\centering
\caption{Benefits of references and revision.}
\label{tab:reference-inspection}
\begin{tabularx}{\linewidth}{>{\hsize=1.5\hsize\raggedright\arraybackslash}X*{2}{>{\hsize=0.75\hsize\centering\arraybackslash}X}}
\toprule
\textbf{Setting}
& \begin{tabular}[c]{@{}c@{}}\textbf{GPT-6}\\\textbf{Astra}\end{tabular}
& \begin{tabular}[c]{@{}c@{}}\textbf{DeepSeek-}\\\textbf{V4-Pro}\end{tabular} \\
\midrule
\rowcolor{analysisband}
\multicolumn{3}{l}{\rule{0pt}{9pt}\textcolor{analysisblue}{\textbf{(a) Reference ablation}}} \\
Without references & 17.3 & 11.1 \\
With references & 52.1 & 37.1 \\
\rowcolor{analysishighlight}
\textbf{Gain} & \textcolor{analysisblue}{\textbf{+34.8}} & \textcolor{analysisblue}{\textbf{+26.0}} \\
\addlinespace[3pt]
\rowcolor{analysisband}
\multicolumn{3}{l}{\rule{0pt}{9pt}\textcolor{analysisblue}{\textbf{(b) Inspection and revision}}} \\
Initial & 47.4 & 30.7 \\
Final & 52.1 & 37.1 \\
\rowcolor{analysishighlight}
\textbf{Gain} & \textcolor{analysisblue}{\textbf{+4.7}} & \textcolor{analysisblue}{\textbf{+6.4}} \\
\bottomrule
\end{tabularx}
\end{minipage}\hspace{0.04\linewidth}%
\begin{minipage}[t]{0.44\linewidth}
\vspace{0pt}
\centering
\caption{Human evaluation agreement.}
\label{tab:human-validation}
\begin{tabularx}{\linewidth}{>{\hsize=1.04\hsize\raggedright\arraybackslash}X*{2}{>{\hsize=0.98\hsize\centering\arraybackslash}X}}
\toprule
\textbf{Category}
& \begin{tabular}[c]{@{}c@{}}{\textbf{Human--}}\\{\textbf{human}}\\[-1pt]{\fontsize{6.5}{7.5}\selectfont Krippendorff $\alpha$}\end{tabular}
& \begin{tabular}[c]{@{}c@{}}{\textbf{Evaluator vs.}}\\{\textbf{expert median}}\\[-1pt]{\fontsize{6.5}{7.5}\selectfont Weighted Cohen $\kappa$}\end{tabular} \\
\midrule
\rowcolor{analysisband}
\multicolumn{3}{l}{\rule{0pt}{9pt}\textcolor{analysisblue}{\textbf{(a) Keypoint-level agreement}}} \\
Visual attributes        & 0.79 & 0.72 \\
Visual style & 0.65 & 0.47 \\
Interaction              & 0.70 & 0.55 \\
\rowcolor{analysishighlight}
\textbf{All keypoints} & \textbf{0.72} & \textbf{0.64} \\
\bottomrule
\end{tabularx}
\par\vspace{3pt}
\begin{tabularx}{\linewidth}{>{\raggedright\arraybackslash}X>{\centering\arraybackslash}p{0.20\linewidth}}
\rowcolor{analysisband}
\multicolumn{2}{l}{\rule{0pt}{9pt}\textcolor{analysisblue}{\textbf{(b) Task-level Pearson $r$}}} \\
Evaluator vs. mean expert score & \textbf{0.74} \\
\bottomrule
\end{tabularx}

\end{minipage}
\end{table}

We further analyze model performance across fine-grained categories, including visual dimensions and requirement types. Additional results are provided in Appendix~\ref{app:complete-results}.

Fig.~\ref{fig:fine-grained-analysis}(a) presents performance across the six visual dimensions defined in \textsection~\ref{sec:benchmark}. Using an equal-weight average across models, palette (92.4), layout (83.1), and components (82.2) score higher than typography (76.1) and overall visual character (58.8). Interaction achieves the lowest score (43.6) and ranks last for nearly all evaluated systems. These results indicate that agents transfer concrete, locally observable attributes more reliably than properties requiring coordination across multiple design choices or interactive states.
Disaggregating keypoints by requirement type reveals differences in difficulty (Fig.~\ref{fig:fine-grained-analysis} (b)). Models perform best on \emph{invert} (98.87) and \emph{specify} (87.68), followed by \emph{copy} (82.42), whereas \emph{adapt} receives the lowest score (75.49). This suggests that selectively transforming a referenced attribute to fit a new context is more challenging than directly reproducing, explicitly specifying, or avoiding it, despite being central to practical reference-guided development.

\subsubsection{Reference Utilization and Visual Inspection Capabilities}
\label{sec:reference-inspection}
As discussed in \textsection~\ref{sec:intro}, effectively incorporating visual references and refining intermediate outputs through visual inspection are important capabilities in agentic generation. We assess these capabilities via our \benchname.
\textbf{(1) Reference ablation.}
We remove all visual references while leaving the original user query unchanged. The agent is explicitly informed that the references mentioned in the query are unavailable and is instructed to satisfy all requirements using its prior knowledge to make reasonable choices.
\textbf{(2) Pre- and post-inspection comparison.}
We evaluate the artifact produced in the first generation round, before the agent receives a rendered view of its output, and compare it with the final artifact.
The results are reported in Tab.~\ref{tab:reference-inspection}. Removing the visual references substantially reduces performance for both models. GPT-6 Astra drops from 52.1 to 17.3 points in overall score, while DeepSeek-V4-Pro drops from 37.1 to 11.1. These results indicate that effectively incorporating the supplied visual references is a crucial capability and that models cannot recover the reference-dependent requirements from the textual query and their priors alone.
We also observe that performance improves from the initial to the final artifacts when models have additional opportunities to inspect and revise their outputs. Although this comparison does not rigorously isolate the causal contribution of visual inspection, since the final artifacts also benefit from additional tool calls, the consistent improvements suggest that the complete inspection-and-revision process is important.

\begin{figure}[t]
\centering
\includegraphics[width=0.90\linewidth]{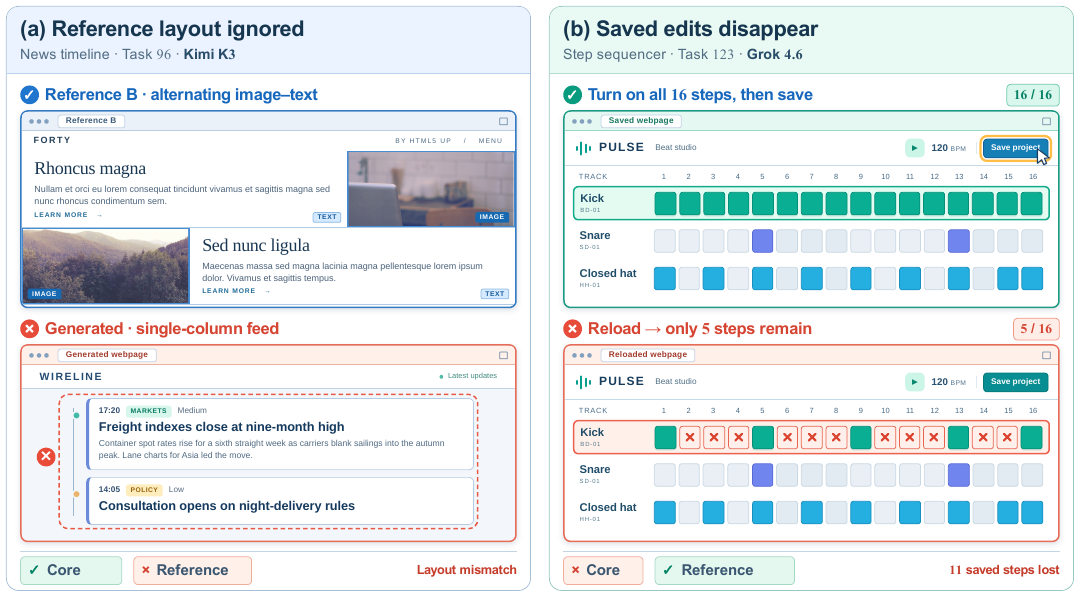}
\caption{Two failure cases identified by \benchname: one in reference following and the other in core functionality.}
\label{fig:human-evaluation-failures}
\end{figure}

\subsubsection{Evaluation Reliability and Qualitative Analysis}
\label{sec:human-validation}
To assess the reliability of our evaluation protocol, we sample 100 model--task artifacts and recruit ten web-development experts to evaluate each artifact independently. We aggregate the expert judgments using the median label for each keypoint and the mean score for each task.
As shown in Tab.~\ref{tab:human-validation}, the evaluator achieves substantial agreement with the expert consensus, both overall and across evaluation aspects, with a weighted $\kappa$ of $0.64$ across all keypoints. More subjective criteria, such as overall visual character, exhibit greater human--human and evaluator--human disagreement, as expected. The task-level scores also correlate strongly with the mean expert scores (Pearson $r=0.747$), indicating that this alignment extends to aggregate assessments of task performance.
We further compare our overall model ranking with the Image-to-Webdev leaderboard from Arena.AI, a popular platform for human evaluation of frontier models\footnote{\url{https://arena.ai/leaderboard/code/image-to-webdev}, accessed September 24, 2026.}. The two rankings achieve a Spearman correlation of $\rho=0.89$, suggesting broad consistency between \benchname and real-world user needs and preferences.

Fig.~\ref{fig:human-evaluation-failures} presents two qualitative failure cases detected by \benchname. In Task~96, Kimi K3 produces a polished news-timeline page that satisfies many keypoints and all core goals. Yet it adopts a centered, single-column feed rather than the requested full-width, two-column image--text layout, thereby failing the reference requirement.
In Task~123, Grok 4.6 produces a step sequencer that follows the visual references, but its save function fails to retain edits, although edit persistence is explicitly requested in the query and can be implemented in JavaScript. Together, these cases illustrate how \benchname identifies both unmet reference requirements and broken core functionality.

\section{Conclusion}

In this paper, we introduced \benchname, a benchmark for studying reference-based webpage development and, more broadly, visual-reference-conditioned agentic generation in frontier models. \benchname comprises 300 realistic web-development tasks in an inspect-and-iterate setting. We decompose each query into independently verifiable keypoints and evaluate the generated artifacts using complementary programmatic and VLM judges. Our evaluation of nine frontier (M)LLMs shows that this capability remains far from solved: the strongest system achieves an overall score of only 52.12. We hope that \benchname will provide a valuable testbed for this crucial yet underexplored capability and stimulate further progress in visual-reference-conditioned agentic generation.

\section*{AI Use Statement}

We used generative AI models in both benchmark construction and evaluation, including synthesizing reference webpages, profiling visual evidence, constructing and reviewing tasks and keypoints, and performing VLM-based evaluation (\textsection~\ref{app:implementation-details}). AI assistants also supported manuscript preparation, including routine tool development and language polishing. The authors reviewed all AI-assisted content and take full responsibility for the final text, experimental claims, and research artifacts.

\section*{Reproducibility Statement}

Task construction, scoring, and implementation details are documented in \textsection~\ref{sec:benchmark}, \textsection~\ref{sec:protocol}, and \textsection~\ref{app:implementation-details}, including prompts, browser settings, and execution budgets. Upon publication, we plan to publicly release \benchname, including task inputs, evaluation scripts, previously generated HTML artifacts, run records, and item-level judgments.

\bibliography{iclr2027_conference}
\bibliographystyle{iclr2027_conference}

\appendix

\newpage
\appendix

\section*{Appendix Overview}

We organize the appendix as follows:
\begin{itemize}[topsep=3pt,itemsep=2pt,parsep=0pt]
    \item Appendix~\ref{app:additional-experiments} presents additional experiments and analyses, including:
    \begin{itemize}[nosep,topsep=2pt]
        \item benchmark statistics (Appendix~\ref{app:benchmark-statistics})
        \item pairwise model comparisons (Appendix~\ref{app:pairwise-comparisons})
        \item judge sensitivity and stability (Appendix~\ref{app:judge-sensitivity})
        \item evaluation protocol ablations (Appendix~\ref{app:protocol-ablation})
        \item complete capability breakdowns (Appendix~\ref{app:complete-results})
        \item cross-harness sensitivity (Appendix~\ref{app:harness-sensitivity})
        \item a worked example of \benchname (Appendix~\ref{app:trajectory-case})
    \end{itemize}
    \item Appendix~\ref{app:implementation-details} provides additional methodological details, including:
    \begin{itemize}[nosep,topsep=2pt]
        \item reference collection and visual evidence (Appendix~\ref{app:reference-pool})
        \item controlled task construction (Appendix~\ref{app:query-generation})
        \item keypoint construction and quality assurance (Appendix~\ref{app:keypoint-validation})
        \item standardized agentic execution (Appendix~\ref{app:agent-implementation})
        \item evidence collection and dual-track judging (Appendix~\ref{app:evaluation-details})
    \end{itemize}
\end{itemize}

\section{Additional Experiments}
\label{app:additional-experiments}

The main experiments establish aggregate model performance and identify challenges in reference utilization and functional execution. Here, we examine how these results vary across individual tasks, and how their interpretation depends on the evaluation and execution setup. These analyses complement the main comparison by distinguishing performance differences from variation introduced by the judge, scoring protocol, or coding harness.

\subsection{Benchmark Statistics}
\label{app:benchmark-statistics}

\begin{figure}[!htbp]
\centering
\includegraphics[width=\textwidth]{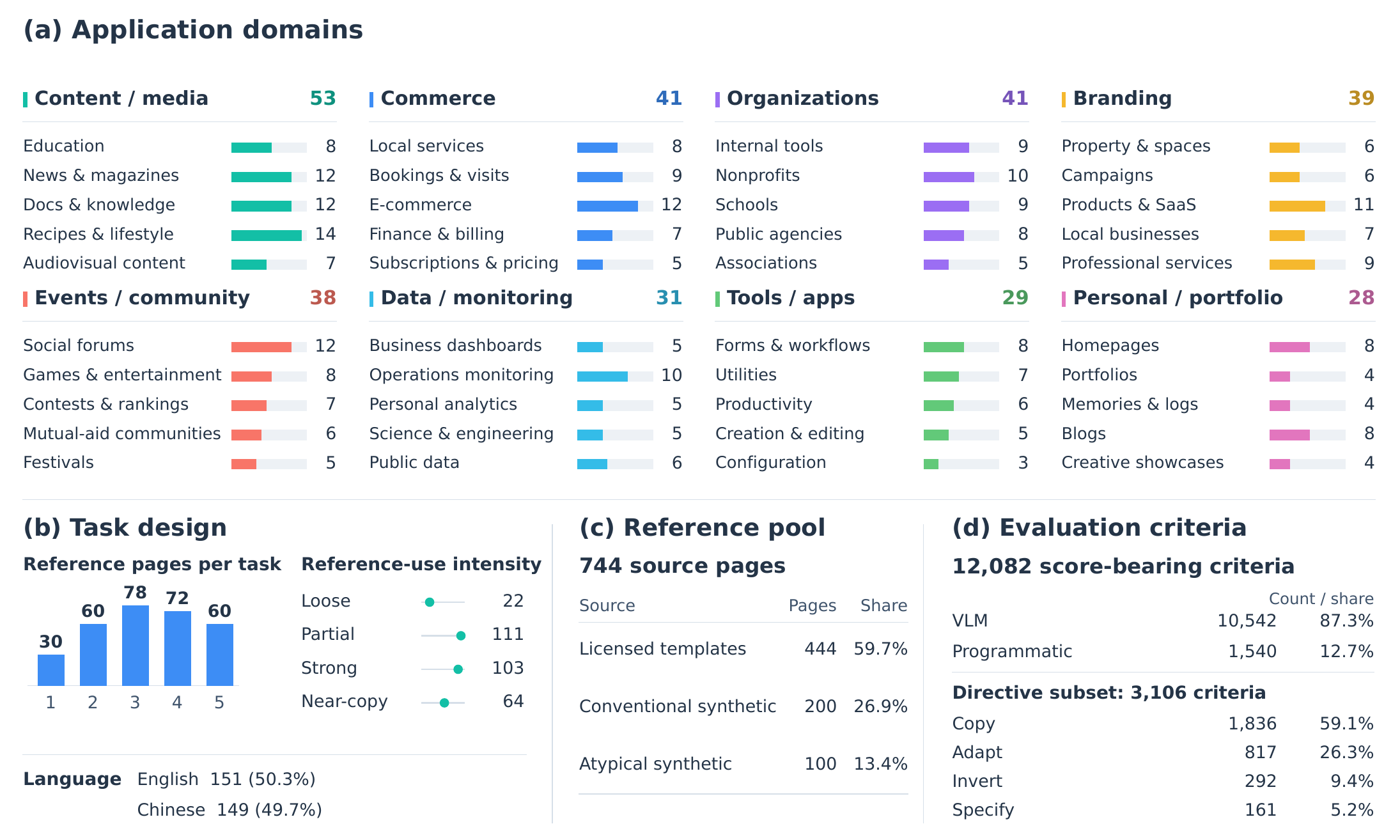}
\caption{Composition and diversity of \benchname.}
\label{fig:benchmark-composition}
\end{figure}

We summarize the composition of \benchname in Fig.~\ref{fig:benchmark-composition}. The final 300 tasks span eight application domains, 40 subdomains, and 186 distinct webpage types. Each task uses one to five reference webpages, with three- and four-reference tasks accounting for 78 and 72 instances, respectively. The benchmark also varies the intended degree of reference use, from loose guidance to near-copying, as determined by requirement types such as \emph{adapt} and \emph{copy}, and includes 151 English and 149 Chinese requests. Together, these distributions capture variation in both what agents are asked to build and how they should use the supplied references. The 744-page reference pool is counted separately because each source page may support multiple tasks.

During evaluation, each keypoint is instantiated as a single evaluation record. This process yields 12,682 records, comprising 10,542 VLM-assessed criteria, 1,540 programmatic probes, and 600 renderability records. The first two categories form the 12,082-item score-bearing set, whereas the renderability records support auditing and core evaluation without contributing additional scoring items. Within the score-bearing set, 3,106 criteria assess directive-related requirements; Fig.~\ref{fig:benchmark-composition}(d) presents their breakdown.

\subsection{Pairwise Model Comparisons}
\label{app:pairwise-comparisons}

We provide detailed task-paired comparisons among the nine evaluated models in Tab.~\ref{tab:pairwise-comparisons}. For each model pair, we compute the mean difference in overall score across the same tasks and report the corresponding proportions of wins, ties, and losses.
The head-to-head results are broadly consistent with the aggregate ranking, while showing that a higher-ranked model does not uniformly outperform its counterpart. In particular, comparisons between closely ranked systems contain substantial proportions of both ties and losses. These paired results clarify how a leaderboard advantage is distributed across tasks, rather than treating a difference in average score as evidence of uniformly superior performance.

\begin{table*}[!htbp]
\centering
\footnotesize
\renewcommand{\arraystretch}{1.08}
\caption{Paired comparisons among the nine main-table models. $\Delta$ is Model A minus Model B.}
\label{tab:pairwise-comparisons}
\rowcolors{2}{tablerow}{white}
\begin{tabularx}{0.96\textwidth}{>{\raggedright\arraybackslash}X>{\raggedright\arraybackslash}Xcccc}
\toprule
\rowcolor{tableheader}
Model A & Model B & $\Delta$ & A win (\%) & Tie (\%) & B win (\%) \\
\midrule
GPT-6 Astra & Qwen3.8-Max & 3.17 & 38.3 & 24.7 & 37.0 \\
GPT-6 Astra & Grok 4.6 & 3.29 & 42.7 & 21.7 & 35.7 \\
GPT-6 Astra & Kimi K3 & 4.57 & 44.0 & 20.7 & 35.3 \\
GPT-6 Astra & Claude Fable 5 & 7.51 & 46.7 & 20.7 & 32.7 \\
GPT-6 Astra & GLM-5.3 & 9.52 & 47.3 & 23.7 & 29.0 \\
GPT-6 Astra & DeepSeek-V4-Pro & 15.03 & 51.0 & 25.3 & 23.7 \\
GPT-6 Astra & Seed2.1 Pro & 20.06 & 61.3 & 22.0 & 16.7 \\
GPT-6 Astra & Gemini 3.1 Pro & 33.11 & 71.0 & 20.3 & 8.7 \\
Qwen3.8-Max & Grok 4.6 & 0.12 & 37.0 & 26.0 & 37.0 \\
Qwen3.8-Max & Kimi K3 & 1.40 & 41.3 & 24.0 & 34.7 \\
Qwen3.8-Max & Claude Fable 5 & 4.34 & 41.0 & 25.0 & 34.0 \\
Qwen3.8-Max & GLM-5.3 & 6.35 & 44.7 & 28.0 & 27.3 \\
Qwen3.8-Max & DeepSeek-V4-Pro & 11.86 & 48.0 & 30.3 & 21.7 \\
Qwen3.8-Max & Seed2.1 Pro & 16.89 & 55.7 & 26.0 & 18.3 \\
Qwen3.8-Max & Gemini 3.1 Pro & 29.94 & 68.0 & 23.7 & 8.3 \\
Grok 4.6 & Kimi K3 & 1.28 & 42.3 & 24.0 & 33.7 \\
Grok 4.6 & Claude Fable 5 & 4.21 & 39.7 & 27.3 & 33.0 \\
Grok 4.6 & GLM-5.3 & 6.23 & 43.3 & 23.3 & 33.3 \\
Grok 4.6 & DeepSeek-V4-Pro & 11.74 & 43.7 & 32.7 & 23.7 \\
Grok 4.6 & Seed2.1 Pro & 16.77 & 54.0 & 28.0 & 18.0 \\
Grok 4.6 & Gemini 3.1 Pro & 29.82 & 67.0 & 26.7 & 6.3 \\
Kimi K3 & Claude Fable 5 & 2.94 & 37.3 & 26.0 & 36.7 \\
Kimi K3 & GLM-5.3 & 4.95 & 41.3 & 26.3 & 32.3 \\
Kimi K3 & DeepSeek-V4-Pro & 10.46 & 44.3 & 32.7 & 23.0 \\
Kimi K3 & Seed2.1 Pro & 15.49 & 53.0 & 26.0 & 21.0 \\
Kimi K3 & Gemini 3.1 Pro & 28.54 & 67.7 & 22.7 & 9.7 \\
Claude Fable 5 & GLM-5.3 & 2.02 & 40.3 & 26.7 & 33.0 \\
Claude Fable 5 & DeepSeek-V4-Pro & 7.52 & 43.3 & 33.0 & 23.7 \\
Claude Fable 5 & Seed2.1 Pro & 12.56 & 49.3 & 31.0 & 19.7 \\
Claude Fable 5 & Gemini 3.1 Pro & 25.60 & 61.3 & 29.0 & 9.7 \\
GLM-5.3 & DeepSeek-V4-Pro & 5.51 & 37.3 & 35.7 & 27.0 \\
GLM-5.3 & Seed2.1 Pro & 10.54 & 47.7 & 29.0 & 23.3 \\
GLM-5.3 & Gemini 3.1 Pro & 23.59 & 60.0 & 27.0 & 13.0 \\
DeepSeek-V4-Pro & Seed2.1 Pro & 5.03 & 37.7 & 39.0 & 23.3 \\
DeepSeek-V4-Pro & Gemini 3.1 Pro & 18.08 & 51.7 & 35.0 & 13.3 \\
Seed2.1 Pro & Gemini 3.1 Pro & 13.05 & 40.7 & 46.0 & 13.3 \\
\bottomrule
\end{tabularx}
\end{table*}

\subsection{Judge Sensitivity and Stability}
\label{app:judge-sensitivity}

\begin{figure}[!htbp]
\centering
\includegraphics[width=0.95\textwidth]{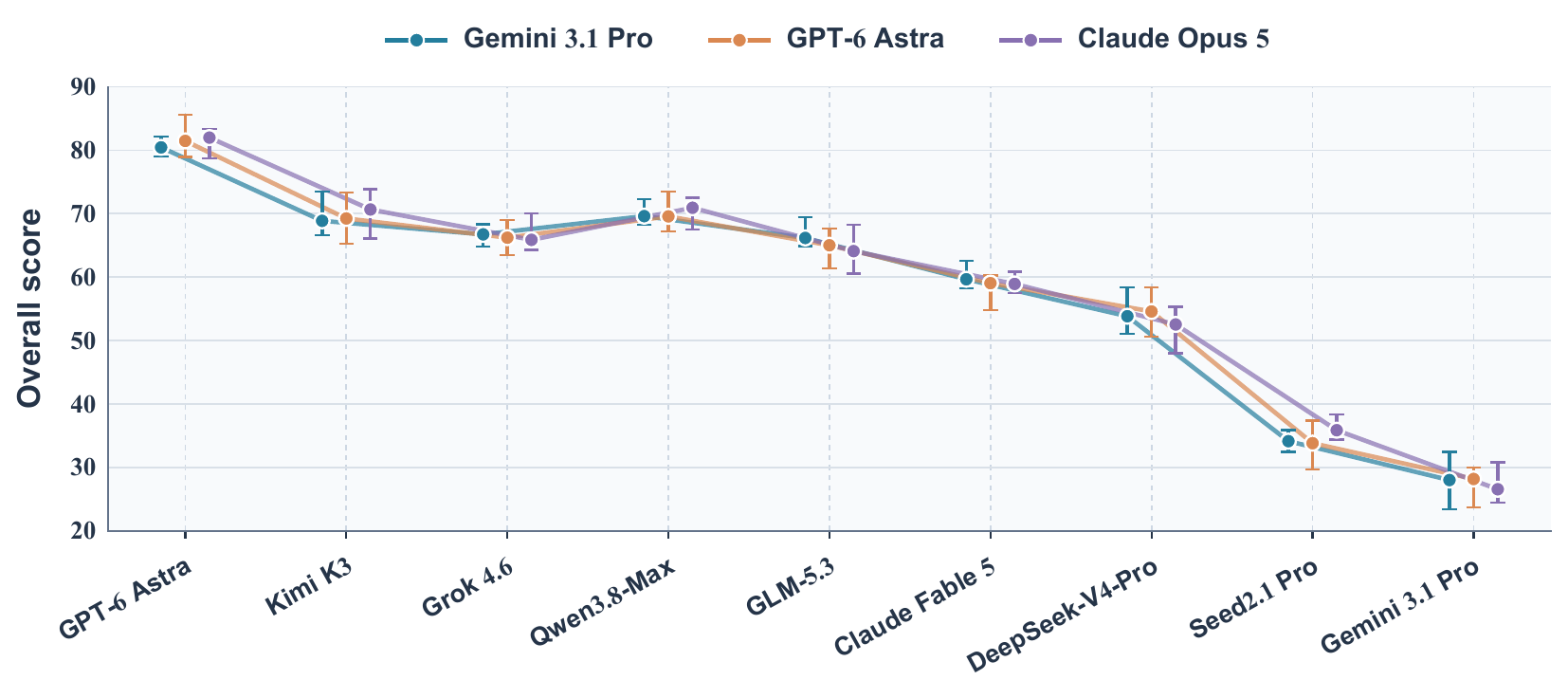}
\caption{Judge sensitivity on the fixed 32-task sample. Colored points and connecting lines show scores assigned by Gemini 3.1 Pro, GPT-6 Astra, and Claude Opus 5. Whiskers indicate illustrative uncertainty intervals. Relative scoring patterns are broadly consistent across the three models.}
\label{fig:app-judge-sensitivity}
\end{figure}

In our evaluation, we use Gemini 3.1 Pro as the VLM judge. A natural question is whether the broad model comparison depends on the choice of automatic evaluator. We examine this sensitivity by re-scoring artifacts from the nine evaluated systems on a fixed 32-task sample using Gemini 3.1 Pro, GPT-6 Astra, and Claude Opus 5. Each judge receives identical tasks, keypoints, reference mappings, rendered views, browser measurements, and interaction evidence. As shown in Fig.~\ref{fig:app-judge-sensitivity}, the three judges produce similar model-level score patterns, suggesting that the broad comparison on this sample is not specific to a single evaluator.

We further examine the stability of the same judge across repeated evaluations. Specifically, we run Gemini 3.1 Pro three times on the same sample with identical inputs. Pairwise correlations between repeated scores reach 0.92, indicating that the overall score pattern is reproducible across runs.

\subsection{Evaluation Protocol Ablations}
\label{app:protocol-ablation}

The protocol in \textsection~\ref{sec:protocol} combines requirement-level assessment with visual and executable evidence. To examine the role of these design choices, we compare the complete protocol with two plausible alternatives on the same generated artifacts. \emph{Holistic evaluation} replaces keypoint decomposition and $B/C/R$ aggregation with a direct overall rating, testing how the assessment changes when requirements are no longer scored and aggregated separately. \emph{Screenshot-only evaluation} retains the rendered images but removes browser measurements and action traces, testing what can be established from visual evidence alone.

\begin{figure}[!htbp]
\centering
\includegraphics[width=\textwidth]{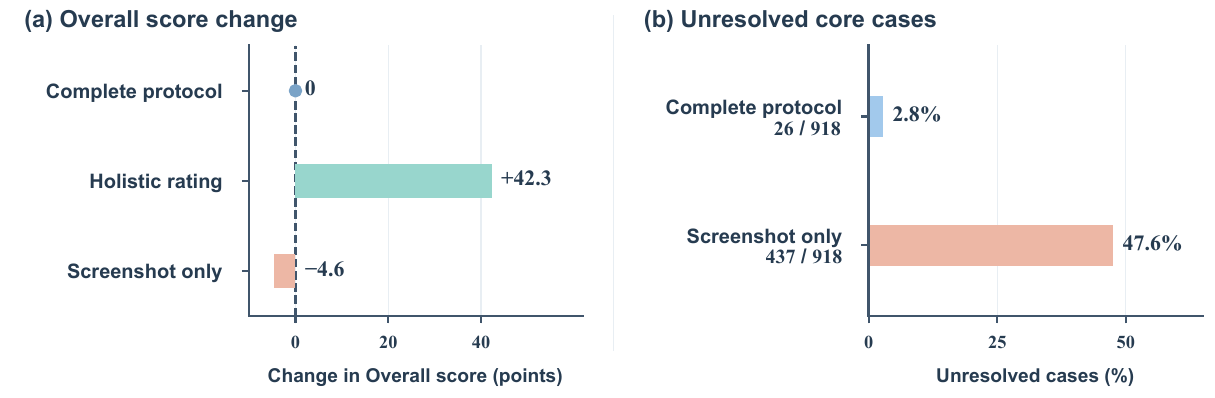}
\caption{Protocol sensitivity and evidence coverage. (a) Changes in Overall score relative to the complete protocol: $+42.3$ under holistic rating and $-4.6$ under screenshot-only evaluation. (b) Observed rates of unresolved core cases across matched artifacts.}
\label{fig:app-protocol-ablation}
\end{figure}

As shown in Fig.~\ref{fig:app-protocol-ablation}, holistic scores exceed complete-protocol scores by 42.3 points. Because holistic evaluation changes both the rating granularity and the aggregation rule, this gap alone cannot establish which protocol better captures artifact quality. Nevertheless, the large increase suggests that holistic evaluation may overestimate model capabilities by overlooking failures exposed by the complete protocol.
Removing executable evidence has a clearer effect on evaluation coverage. Unresolved core cases increase from 26 to 437 out of 918, while unresolved keypoints increase from 15 to 1,987 out of 7,608. These results indicate that visual evidence alone is often insufficient to determine whether requirements are satisfied, making screenshot-only evaluation incomplete and unreliable for assessing overall artifact behavior.

\subsection{Complete Capability Breakdowns}
\label{app:complete-results}

The fine-grained analysis in \textsection~\ref{sec:fine-grained-analysis} focuses on visual dimensions and requirement types. Tab.~\ref{tab:domain-breakdown} and~\ref{tab:source-breakdown} complement this analysis by examining application domains and evidence sources across the nine evaluated models. The domain breakdown reports task-level overall scores, capturing end-to-end performance within each task group. The source breakdown instead reports reference-use keypoint scores, obtained by first aggregating the relevant keypoints within each task and then averaging across tasks. Across domains, equal-model averages are highest for personal and portfolio (44.7), organizations (44.4), and commerce (44.1), and lowest for data and monitoring (33.7) and events and community (37.1).
Tab.~\ref{tab:source-breakdown} shows that models score higher on text-specified requirements than on requirements associated with atypical synthetic, licensed/template, or conventional synthetic references. This gap suggests that explicit textual constraints are easier to follow than properties that must be inferred from visual references.

\begin{table*}[!htbp]
\centering
\small
\setlength{\tabcolsep}{5pt}
\renewcommand{\arraystretch}{1.13}
\caption{Reference-related keypoint performance by evidence source.}
\label{tab:source-breakdown}
\begin{adjustbox}{max width=\textwidth}
\rowcolors{2}{tablerow}{white}
\begin{tabular}{lcccc}
\toprule
\rowcolor{tableheader}
Model & \ensuremath{\vcenter{\hbox{\shortstack{Licensed/\\template}}}} & \ensuremath{\vcenter{\hbox{\shortstack{Atypical\\synthetic}}}} & \ensuremath{\vcenter{\hbox{\shortstack{Conventional\\synthetic}}}} & \ensuremath{\vcenter{\hbox{\shortstack{Text\\specification}}}} \\
\midrule
GPT-6 Astra & 82.7 & 78.8 & 80.7 & 91.6 \\
Qwen3.8-Max & 81.9 & 80.0 & 88.5 & 90.9 \\
Grok 4.6 & 83.3 & 82.5 & 87.4 & 91.1 \\
Kimi K3 & 81.6 & 79.3 & 81.7 & 85.5 \\
Claude Fable 5 & 81.7 & 78.5 & 88.0 & 87.4 \\
GLM-5.3 & 80.5 & 75.7 & 80.8 & 88.1 \\
DeepSeek-V4-Pro & 79.3 & 76.0 & 80.3 & 86.1 \\
Seed2.1 Pro & 80.0 & 76.8 & 81.9 & 81.8 \\
Gemini 3.1 Pro & 80.9 & 81.7 & 83.7 & 86.6 \\
\bottomrule
\end{tabular}%
\end{adjustbox}
\end{table*}

\begin{table*}[!htbp]
\centering
\footnotesize
\setlength{\tabcolsep}{3pt}
\renewcommand{\arraystretch}{1.13}
\caption{Task-level Overall score across the eight high-level application domains.}
\label{tab:domain-breakdown}
\begin{adjustbox}{max width=0.96\textwidth}
\rowcolors{2}{tablerow}{white}
\begin{tabular}{l*{8}{c}}
\toprule
\rowcolor{tableheader}
Model & \ensuremath{\vcenter{\hbox{\shortstack{Content/\\media}}}} & Commerce & Organizations & Branding & \ensuremath{\vcenter{\hbox{\shortstack{Events/\\community}}}} & \ensuremath{\vcenter{\hbox{\shortstack{Data/\\monitoring}}}} & \ensuremath{\vcenter{\hbox{\shortstack{Tools/\\apps}}}} & \ensuremath{\vcenter{\hbox{\shortstack{Personal/\\portfolio}}}} \\
\midrule
GPT-6 Astra & 50.2 & 57.6 & 53.9 & 54.4 & 46.8 & 46.0 & 54.4 & 53.6 \\
Qwen3.8-Max & 49.2 & 50.5 & 50.2 & 49.4 & 42.2 & 50.0 & 49.4 & 51.2 \\
Grok 4.6 & 46.0 & 52.0 & 55.6 & 49.3 & 49.6 & 34.7 & 49.6 & 52.6 \\
Kimi K3 & 43.9 & 49.4 & 53.5 & 46.8 & 39.5 & 44.5 & 51.2 & 54.7 \\
Claude Fable 5 & 42.6 & 50.6 & 50.7 & 43.8 & 43.5 & 35.8 & 46.9 & 40.9 \\
GLM-5.3 & 45.9 & 40.4 & 45.8 & 39.9 & 38.2 & 35.7 & 48.2 & 46.7 \\
DeepSeek-V4-Pro & 40.3 & 38.7 & 37.7 & 38.2 & 32.4 & 25.0 & 37.4 & 45.6 \\
Seed2.1 Pro & 33.9 & 37.9 & 32.2 & 38.8 & 22.0 & 20.2 & 33.0 & 36.4 \\
Gemini 3.1 Pro & 16.1 & 19.8 & 20.4 & 20.3 & 19.9 & 11.3 & 24.7 & 21.0 \\
\bottomrule
\end{tabular}%
\end{adjustbox}
\end{table*}

\subsection{Cross-Harness Sensitivity}
\label{app:harness-sensitivity}

The shared \texttt{OpenCode} configuration ensures a consistent main comparison across models, but real-world agent systems use diverse harnesses. To test sensitivity to the choice of harness, we evaluate GPT-6 Astra and Kimi K3 on a fixed 32-task sample using two popular alternatives: Claude Code\footnote{\url{https://github.com/anthropics/claude-code}} and Codex\footnote{\url{https://github.com/openai/codex}}. We also include the released OpenCode artifacts for the same 32 tasks, re-scored using the same evaluation protocol, as the main-harness baseline.

For the Claude Code and Codex reruns, we match the task text, ordered references, initial workspaces, and nominal resource limits, while retaining each harness's native system prompts, tools, image preprocessing, and context-management mechanisms. Tab.~\ref{tab:cross-harness-sensitivity} summarizes the overall scores for each model--harness combination.
Scores vary modestly across harnesses. GPT-6 Astra ranges from 49.2 to 53.0, a spread of 2.8 points, while Kimi K3 ranges from 43.1 to 49.4, a spread of 1.1 points. GPT-6 Astra outperforms Kimi K3 under all three harnesses, preserving their relative ordering. These results suggest that the harness affects absolute scores, while overall performance and model ordering remain relatively stable on this sample.

\begin{table}[!htbp]
\centering
\small
\caption{Cross-harness sensitivity on the fixed 32-task sample. We report Overall scores across three harnesses.}
\label{tab:cross-harness-sensitivity}
\renewcommand{\arraystretch}{1.13}
\begin{tabularx}{0.85\textwidth}{>{\raggedright\arraybackslash}X>{\raggedright\arraybackslash}Xc}
\toprule
\rowcolor{tableheader}
Model & Harness & Overall score \\
\midrule
GPT-6 Astra & OpenCode (released) & $49.2$ \\
GPT-6 Astra & Claude Code & $51.3$ \\
GPT-6 Astra & Codex       & $53.0$ \\
\midrule
Kimi K3     & OpenCode (released) & $48.2$ \\
Kimi K3     & Claude Code & $49.4$ \\
Kimi K3     & Codex       & $43.1$ \\
\bottomrule
\end{tabularx}
\end{table}

\subsection{A Worked Example of \benchname}
\label{app:trajectory-case}

\begin{figure}[H]
\centering
\includegraphics[width=\textwidth,height=0.84\textheight,keepaspectratio]{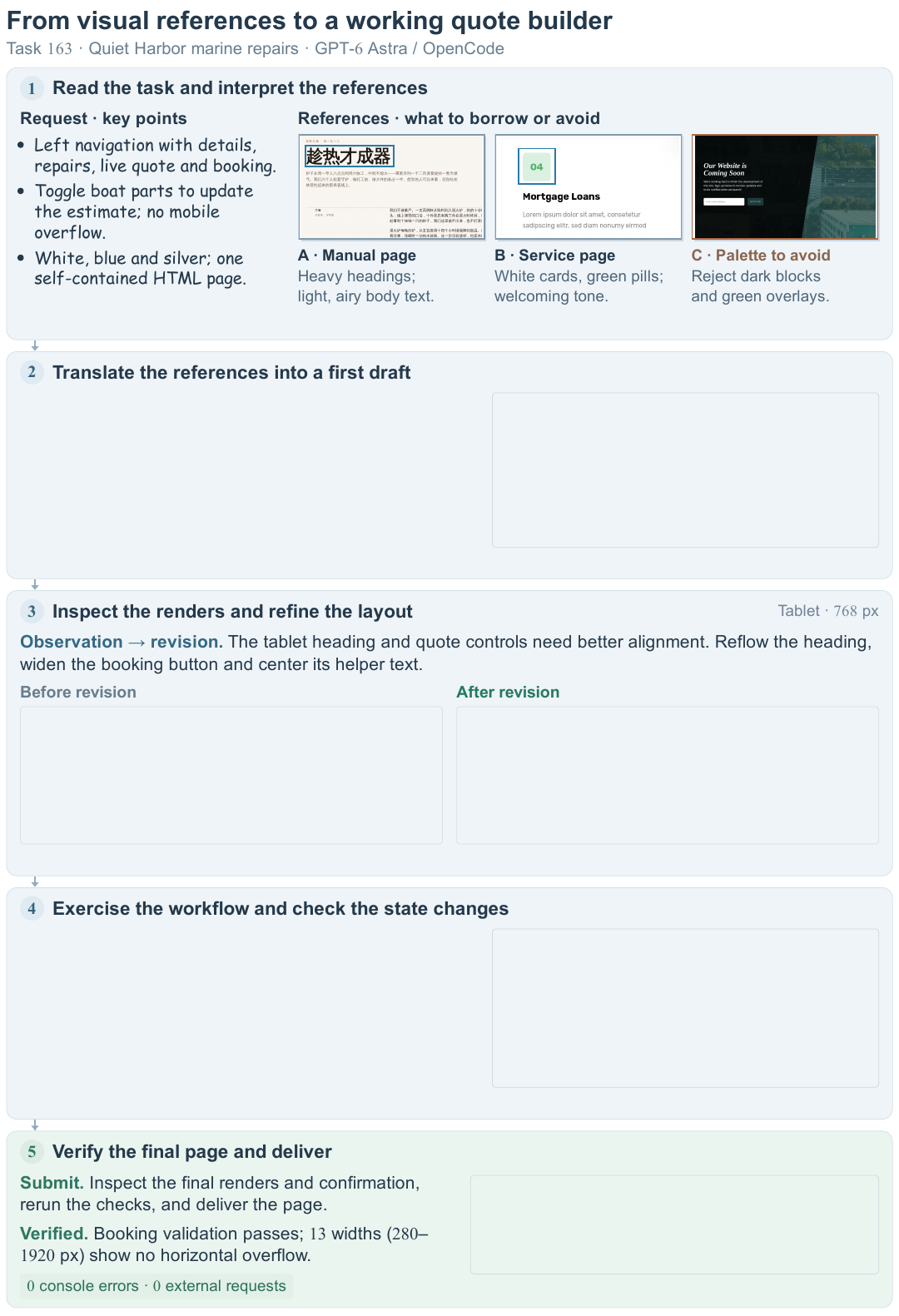}
\caption{An end-to-end development trajectory on Task~163. Boxes and enlarged screenshot crops highlight reference cues, responsive-layout revisions, and changes in the quote and booking state.}
\label{fig:trajectory-workflow}
\end{figure}

Fig.~\ref{fig:trajectory-workflow} traces how GPT-6 Astra, running in OpenCode, completes Task~163: a marine repair quote builder. The model first identifies which properties to borrow from each reference and which to avoid, combining bold typography and light service cards while rejecting the dark reference palette. It then builds a self-contained \texttt{index.html}, inspects rendered views, and revises the tablet heading and booking controls. Functional checks verify that repair selections update the quote and that editing a confirmed plan reopens the booking form; the final page also shows no horizontal overflow across 13 tested widths. This example illustrates how reference interpretation, visual refinement, and interaction testing work together to turn a user request into a functional webpage.

\section{Additional Implementation Details}
\label{app:implementation-details}

\textsection~\ref{sec:benchmark} and \textsection~\ref{sec:protocol} describe the design of \benchname and its evaluation protocol. Here, we explain how these designs are implemented. We follow the same progression as the main text: reference webpages are first converted into visual evidence, which supports the construction of user requests and their evaluation keypoints. Once these task components are fixed, an agent uses the request and reference images to develop a webpage, while a separate evaluator assesses the submitted artifact against the hidden keypoints. The details below apply to the \benchname benchmark and clarify how information passes between these stages.

\subsection{Reference Collection and Visual Evidence}
\label{app:reference-pool}

We begin by preparing a reusable collection of webpages from which task-specific references can be assembled. Following the three-source design in \textsection~\ref{sec:reference-pool}, this collection contains 744 pages: 444 openly licensed template-derived pages, 200 conventionally styled synthetic pages, and 100 atypical synthetic pages. These counts describe the available source pool rather than the benchmark instances. A task draws visual references from one to five source pages, and the same source page may support different requests that ask agents to use its properties in different ways.

\paragraph{Template sources and licensing.}
For the template-derived portion, we collect openly licensed pages from collections including
\href{https://html5up.net/}{HTML5 UP},
\href{https://github.com/ColorlibHQ/bootstrap-templates}{Colorlib},
\href{https://startbootstrap.com/}{Start Bootstrap},
\href{https://www.creative-tim.com/}{Creative Tim},
\href{https://github.com/ColorlibHQ/AdminLTE}{AdminLTE},
\href{https://github.com/themefisher}{Themefisher}, and
\href{https://github.com/codewithsadee}{CodeWithSadee}. The retained copies comprise 367 MIT-licensed pages, 76 under CC BY 3.0, and one under Apache-2.0. For each page, we record its source URL, author, license, and design-family identifier, preserving the connection between the benchmark reference and its original source. We also package the page's styles, scripts, and image assets into a self-contained HTML snapshot. This snapshot provides a stable input to the capture process: the browser renders the stored page rather than relying on the continued availability or current appearance of the live website.

\paragraph{Synthetic reference generation.}
The synthetic portion complements these existing templates by extending the range of designs available for task construction. Across the 300 synthetic pages, we use GPT-5.2 and GPT-5.4~\citep{openai2025gpt52,openai2026gpt54}, Claude Opus 4.5, Claude Opus 4.6, Claude Opus 4.8, and Claude Sonnet 4.6~\citep{anthropic2025opus45,anthropic2026opus46,anthropic2026opus48,anthropic2026sonnet46}, Gemini 3.1 Pro~\citep{google2026gemini31pro}, Qwen3.6-Plus~\citep{qwen2026qwen36}, Seed2.0 Pro~\citep{bytedance2026seed20}, GLM-5~\citep{glm5team2026glm5}, and MiniMax M2.7~\citep{minimax2026m27} as reference generators. As discussed in the main text, however, model-generated pages can themselves converge on familiar design patterns. For the 100 atypical pages, we therefore pair a controlled design sampler with Claude Opus 4.8~\citep{anthropic2026opus48} to introduce less conventional visual combinations. The sampler combines 16 predefined design families (e.g., Swiss typography~\footnote{\url{https://en.wikipedia.org/wiki/Swiss_Style_(design)}}, brutalism, and terminal interfaces) with varied palettes and family-specific font stacks, using flat fills, square corners, and optional hard shadows. The model then refines the copy and page structure, while automated checks enforce the palette, font, and styling constraints. These models participate in constructing the reference pool, a role distinct from that of the nine systems subsequently evaluated on the benchmark.

\paragraph{From webpages to task evidence.}
The stored webpages are construction assets, but the references shown to agents consist only of images. To produce these visual inputs, Playwright renders each page at a $1920\times1080$ viewport and captures scripted scrolling and interaction states, yielding 5,522 frames across the pool. For conventionally styled synthetic pages, we retain their original page-specific demonstration scripts. For template-derived and atypical pages, we generate scripts from the measured page height and available hover targets (identified through known selectors or DOM probing), interleaving stepwise scrolling, hovers, and short pauses. We retain the capture-action script and a time-stamped frame manifest for each page so that the resulting images remain associated with the actions and source that produced them. When references are assembled into a task, frame order and source-page boundaries are preserved. The agent can thus distinguish views of the same page from views of different references without receiving their implementations.
Before a request can be written, we also need to establish which reference properties these images actually expose. Gemini 3.1 Pro~\citep{google2026gemini31pro}
profiles the frames along the six visual dimensions in \textsection~\ref{sec:reference-pool}, providing a description of the visible evidence and a \texttt{rich}, \texttt{weak}, or \texttt{none} label for each dimension. These profiles connect reference collection to query synthesis: they restrict the assignment of reference-dependent requirements to properties supported by the supplied images, rather than by information available only in the underlying webpage.

\subsection{Controlled Task Construction}
\label{app:query-generation}

We construct each user request from a structured recipe that combines the reference-use specification in \textsection~\ref{sec:query-synthesis} with choices about the target webpage and how the user expresses their needs. The target domain is drawn from a taxonomy of 8 broad categories, 40 subcategories, and 206 site types, ranging from productivity tools such as kanban boards to salon booking pages and API documentation. The recipe also selects one of 14 user profiles, each defining an identity, priorities, and speaking style. For example, a shop owner may emphasize contact details in everyday language, whereas a front-end engineer focuses on semantic structure and responsive behavior.

The recipe also controls the intended page's functional complexity, content density, and explicit constraints. Functional complexity ranges from static pages to stateful applications and data visualization. Up to three tags specify requirements concerning audience, presentation, content volume, delivery, tone, or functionality. These can call for mobile-first display, long-form text, image-led content, offline operation, or bilingual switching. Concrete visual requirements, such as body text of at least 18px, a specified brand color, or square-cornered buttons, are recorded independently of the references.

Separately, we sample the language, form, and level of detail of the request. Requests are written in Chinese or English and take the form of flowing prose, casual messages, a flat list of requirements, or a structured product requirements document (PRD) with section headings and acceptance criteria. The level of detail is sampled independently of the page's functional complexity. It ranges from a single overall need, through a brief description of three to five sections, to detailed content, data, and interactions for each section. Target length depends on both form and level of detail and is measured in Chinese characters or English words. Two exemplars matching the request's language and form guide organization and tone without supplying topics or wording to reuse.

\begin{figure}[!htbp]
\centering
\includegraphics[width=0.85\linewidth]{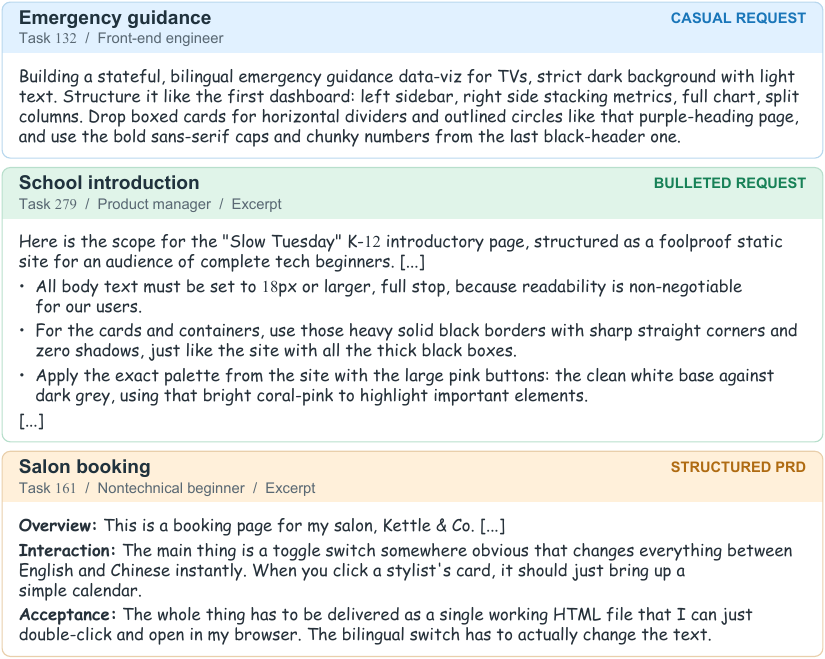}
\caption{Generated user query examples. English requests from the final task set illustrate casual, bulleted, and structured PRD forms. Task 132 is shown in full; Tasks 279 and 161 are excerpted, with \texttt{[...]} marking omissions. Original wording is retained.}
\label{fig:generated-query-examples}
\end{figure}

Gemini 3.1 Pro then converts the recipe and ordered screenshot groups into a natural-language request using the prompt in Fig.~\ref{fig:implementation-prompts}. The prompt asks the model to identify each source by its appearance or position and explain how its selected properties should inform the design. Each reference is used only for the specified aspects of the page, and its content is not reused. Explicit values, such as the 18px minimum for body text, must be preserved so that constraints remain checkable after rewording. Visual descriptions must be grounded in the screenshots; motion that is not visible in them is phrased as a new user requirement. We retain the resulting request with its recipe for keypoint construction. Fig.~\ref{fig:generated-query-examples} shows three examples with different forms and user profiles.

\subsection{Keypoint Construction and Quality Assurance}
\label{app:keypoint-validation}

Keypoint construction must preserve the reference relationships established during query synthesis while also covering the content and functionality expressed in the completed request. We handle these two parts separately. Reference assignments and explicit textual constraints are converted programmatically into fixed items from the structured specification, retaining their intended sources and requirement types. Claude Opus 5~\citep{anthropicopus5} then reads the user request and adds keypoints for content, structure, navigation, and functionality without rewriting the fixed items. This division keeps the specified reference relationships intact while allowing the keypoints to cover requirements that emerge in the natural-language request.

To make the added keypoints assessable, the construction prompt (Fig.~\ref{fig:implementation-prompt-checklist}) links each criterion to both the user request and the evidence needed to evaluate it. Each item must first cite the supporting request text and clarify what it checks: the existence of an element, the correctness of its implementation, or the precision of its details. Distinguishing these levels prevents the mere presence of an element from being treated as sufficient evidence that it has been implemented as instructed. The resulting criterion must then be answerable from the evidence available during evaluation, ruling out questions requiring source-code inspection or unavailable timing measurements. For interaction-dependent requirements, this also means identifying both the control and the action needed to reach the relevant state, so that the subsequent browser-capture process can expose the behavior being assessed.

The generated keypoints are then reviewed by GPT-5.6 Sol~\citep{openai2026gpt56sol} with high reasoning effort and the prompt shown in Fig.~\ref{fig:implementation-prompt-review}. The review separates completeness from item validity: a global pass first proposes missing requirements, after which a local pass examines individual items and returns \texttt{keep}, \texttt{modify}, or \texttt{remove}, together with a reason. Both passes use the reference-use specification to check whether the criteria preserve the intended task. In particular, an ignored dimension must not introduce a requirement, an adaptation must not become an exact-match demand, and an inversion must not reward copying. These checks address errors that could otherwise make a correct implementation appear unsuccessful. Generation and model review use a configured maximum output of 128,000 tokens. Finally, the expert review described in \textsection~\ref{sec:benchmark} revises the keypoints before they are used for evaluation. Fig.~\ref{fig:expert-review-instructions} presents the instruction template for expert keypoint review.

\paragraph{Final task set.}
The final benchmark contains 300 tasks,  each pairing a user request and ordered reference screenshots with a verified set of keypoints. Together, these tasks use 534 distinct webpages in 972 reference assignments and expose 3,491 reference frames. Stable task identifiers and input hashes preserve the correspondence between each request, its references, and its evaluation specification when these components are passed to the execution and judging stages. 

\subsection{Standardized Agentic Execution}
\label{app:agent-implementation}

With the task components fixed, execution exposes only the information intended for the evaluated agent: the user request, ordered reference images, and an environment brief. The verified keypoints remain hidden and are used later by the evaluator. All released runs use OpenCode 1.18.25 with the same 100-turn cap, implementing the shared coding-agent setting described in \textsection~\ref{sec:evaluation-environment}.

The environment brief is matched to the language of the request and explains both the required deliverable and the available development workflow. It asks the agent to ``write the finished page to \texttt{index.html},'' requires a single HTML file, and states that ``grading renders the page offline.'' It also describes the inspection loop explicitly as ``draft $\rightarrow$ screenshot $\rightarrow$ compare $\rightarrow$ fix $\rightarrow$ screenshot again.'' To make this loop executable, the workspace provides shell and file-editing tools together with a Playwright-based \texttt{shot.py} helper. The helper renders the current HTML at widths of 1920, 768, and 390 pixels and returns image paths along with page height, console errors, and attempted external requests. These outputs allow the agent to inspect its rendered implementation and identify problems before revising the page.

Network access remains available during development, but it does not relax the offline submission requirement. Dependencies must be inlined in the final page so that its evaluation does not rely on external resources. Likewise, the availability of development tools does not grant access to benchmark data, hidden criteria, or original reference implementations. Such access is denied and audited. At completion, we retain one official artifact per model--task pair, together with a run record containing the enforced limits, stop reason, transport attempts, token usage, and artifact hashes. This record links the webpage evaluated downstream to the trajectory and resource use reported for that run.

\subsection{Evidence Collection and Dual-Track Judging}
\label{app:evaluation-details}

The agent's inspection tools support development, but the evidence used for scoring is collected separately from the submitted \texttt{index.html}. This distinction ensures that the evaluator examines the final webpage through its own capture and measurement procedure rather than relying on the views the agent happened to inspect. The hidden keypoints guide this procedure by identifying the states and properties that need to be exposed. The resulting evidence is then passed to the programmatic and VLM judging tracks described in \textsection~\ref{sec:scoring}.

\paragraph{Capture and measurement settings.}
Playwright/Chromium first renders the artifact at $1920\times1080$, with a target of five to ten distinct primary frames. Since requirements can concern states that are not visible on initial load, a deterministic action planner uses the keypoints to plan at most eight states to expose. For each item, it tries up to four candidate controls and allows 900\,ms for a state transition. These settings govern how the browser attempts to obtain the interaction evidence specified during keypoint construction.

The evaluator also measures the page at widths of 1280, 768, and 390 pixels to assess behavior beyond the primary desktop view. Additional stress fixtures test robustness when content differs from the initial presentation: eligible collections are replaced with zero or 300 rows, an 80-character unbroken string is injected, and mixed CJK/emoji content is supplied. The resulting measurements complement the ordinary rendered views by exposing behavior under narrow layouts and abnormal content. If a measurement is unavailable or inapplicable, that status is recorded explicitly. The absence of a usable measurement is not treated as evidence that the corresponding test succeeded.

\paragraph{From evidence to judgments.}
For program-verifiable keypoints, the collected browser measurements determine pass/fail verdicts. Requirements that need semantic or perceptual interpretation are instead assigned to Gemini 3.1 pro, configured with high thinking level and a maximum output of 65,535 tokens. This judge receives visual and measured evidence rather than the submitted HTML, reference source code, or solver identity. Its task is to assess what the webpage demonstrates against the stated requirement, rather than infer success from implementation claims.

The perceptual judge processes batches of 5 keypoints, with a separate assessment for each item. Each batch contains the user request, the supporting basis for each item, the reference assignments, and normalized browser facts. To keep these facts and images interpretable, the images follow a fixed order: reference frames, up to ten main artifact views, up to two views per narrow viewport, and up to two stress-fixture views. An accompanying manifest identifies the groups, making clear which images describe the reference and which show the submitted page under each evaluation condition.

Before assigning a rating, the judge must record the \texttt{expected} requirement, the \texttt{observed} evidence, and the \texttt{deviation} between them. This sequence asks it to establish a concrete comparison before choosing a score. Fig.~\ref{fig:implementation-prompt-judge} presents these audit instructions and the rating rubric. The full template also specifies evidence handling and exceptions for unavailable external inputs. The resulting item scores are combined with core-function evidence using the aggregation rule in \textsection~\ref{sec:scoring}. We retain render manifests, measurements, action outcomes, and item verdicts separately, allowing the recorded judgments to be inspected and the scores to be re-aggregated without another model call.

\begin{figure}[!htbp]
\centering
\input{prompts/implementation_task_panel}
\caption{Prompt for user-query generation. Across all prompt figures, \texttt{[...]} denotes omitted passages. Formatting and line breaks are adjusted for presentation.}
\label{fig:implementation-prompts}
\end{figure}

\begin{figure}[!htbp]
\centering
\definecolor{promptbody}{HTML}{F8FCFF}
\definecolor{promptheader}{HTML}{DDEFFF}
\definecolor{promptborder}{HTML}{86BCEC}
\definecolor{prompttitle}{HTML}{184D79}
\begin{tcolorbox}[
    enhanced,
    colback=promptbody,
    colframe=promptborder,
    colbacktitle=promptheader,
    coltitle=prompttitle,
    fonttitle=\bfseries,
    title={Expert keypoint review},
    boxrule=0.5pt,
    arc=1.5mm,
    left=9pt,right=9pt,top=7pt,bottom=7pt,
    toptitle=5pt,bottomtitle=5pt
]
\small
\setlength{\parindent}{0pt}
\setlength{\parskip}{4pt}

\textbf{Your role.} Review and revise the evaluation checklist so that it faithfully represents the user's request and can be applied to different valid implementations. You are reviewing the evaluation criteria, not scoring a generated webpage. Do not use model outputs or scores to decide which requirements belong in the checklist.

\textbf{Materials.} Read the user instruction, ordered reference screenshots, reference-use specification, and draft checklist, including each item's identifier, supporting basis, and annotations. Review both seeded and model-generated items.

\begin{enumerate}[leftmargin=*,label=\textbf{\arabic*.},topsep=4pt,itemsep=5pt,parsep=0pt]
    \item \textbf{Check coverage against the task.}
    Identify the requested content, structure, navigation, visual properties, and essential behavior, then check whether the checklist covers them. Add missing requirements with a supporting basis. If a function is implicit but necessary to fulfill the stated task, explain that necessity explicitly. Do not add personal design preferences or optional conventions.

    \item \textbf{Ground each item in the correct evidence.}
    Quote the relevant request text and, for a reference-dependent property, identify the source page and frame supporting it. Confirm that the cited feature is visible and belongs to that reference. Remove invented details and unsupported requirements. If the request, screenshots, and specification conflict, flag the inconsistency rather than silently changing the task.

    \item \textbf{Preserve the intended reference relationship.}
    Check each constrained dimension against its assigned requirement type:
    \begin{itemize}[leftmargin=1.2em,topsep=3pt,itemsep=1pt,parsep=0pt]
        \item \texttt{copy}: assess the requested property against the assigned reference.
        \item \texttt{adapt}: assess the selected characteristics and requested transformation, without adding exact-match demands.
        \item \texttt{invert}: assess the requested departure from the reference, not similarity to it.
        \item \texttt{specify}: follow the explicit textual constraint rather than importing a reference property.
        \item \texttt{ignore}: do not introduce a requirement for the unconstrained dimension.
    \end{itemize}

    \item \textbf{Make the criteria specific and verifiable.}
    Each item should assess one requirement without duplicating another. Distinguish existence, correctness, and precision, and do not invent numerical tolerances. For interactions, name the control, action, and observable outcome. Check that the criterion can be judged from rendered views or available browser measurements, not source-code claims or unseen behavior. Use programmatic checks where reliable measurement is available and perceptual judgment otherwise. Flag incorrect source, dimension, depth, importance, or judging-track annotations.

    \item \textbf{Return an auditable revision.}
    Mark each existing item as \texttt{keep}, \texttt{modify}, or \texttt{remove}, and label new items \texttt{add}. Preserve existing identifiers. For every change or addition, provide the revised wording where applicable, supporting evidence, corrected annotations, and a brief reason. Return the revised checklist with a separate list of unresolved ambiguities or missing evidence. Do not approve unresolved items by assumption or change the scoring rubric and aggregation rule.
\end{enumerate}
\end{tcolorbox}
\caption{Instruction template for expert keypoint review. Reviewers check coverage, evidence grounding, reference-use consistency, and verifiability, then return revised keypoints with reasons for changes and unresolved issues.}
\label{fig:expert-review-instructions}
\end{figure}

\begin{figure}[!htbp]
\centering
\input{prompts/implementation_checklist_panel}
\caption{Keypoint-construction prompt excerpts.}
\label{fig:implementation-prompt-checklist}
\end{figure}

\begin{figure}[!htbp]
\centering
\input{prompts/implementation_review_panel}
\caption{Keypoint-review prompt excerpts. The global pass checks coverage, and the local pass reviews individual items against the reference-use specification and supporting evidence. }
\label{fig:implementation-prompt-review}
\end{figure}

\begin{figure}[!htbp]
\centering
\definecolor{promptbody}{HTML}{F8FCFF}
\definecolor{promptheader}{HTML}{DDEFFF}
\definecolor{promptborder}{HTML}{86BCEC}
\definecolor{prompttitle}{HTML}{184D79}
\begin{tcolorbox}[
    enhanced,
    colback=promptbody,
    colframe=promptborder,
    colbacktitle=promptheader,
    coltitle=prompttitle,
    fonttitle=\bfseries,
    title={Perceptual-judging prompt},
    boxrule=0.5pt,arc=1.5mm,
    left=9pt,right=9pt,top=7pt,bottom=7pt,
    toptitle=5pt,bottomtitle=5pt
]
\fontsize{9}{9.5}\selectfont
\setlength{\parindent}{0pt}
\setlength{\parskip}{1.25pt}
\par
{\color{black!45}\texttt{[...]}}

\par
For each item, determine the following three things in order and include them in the output:

\begin{enumerate}[leftmargin=1.4em,label=\arabic*.,start=1,topsep=1pt,itemsep=1.5pt,parsep=0pt]
\item[1.]
\textbf{expected (required value)}: Extract a \textbf{specific, verifiable requirement} from {\mdseries\ttfamily basis}. Write a value, not a paraphrase.
\begin{itemize}[leftmargin=1.2em,topsep=1pt,itemsep=1pt,parsep=0pt]
\item
\ensuremath{\checkmark} {\rmfamily "5 cards"} / {\rmfamily "dominant color \#1a2340 (from the reference screenshot header)"} / {\rmfamily "exact text: 'Live leaderboard'"}

\item
\ensuremath{\times} {\rmfamily "Colors must match the reference"} (this restates the requirement rather than giving a value).

\end{itemize}

\item[2.]
\textbf{observed (measured value)}: The \textbf{actually measured value of the same dimension} in the \textbf{fact block}/rendered frames. This must also be a value.
\begin{itemize}[leftmargin=1.2em,topsep=1pt,itemsep=1pt,parsep=0pt]
\item
\ensuremath{\checkmark} {\rmfamily "3 cards (see frame 04)"} / {\rmfamily "\#4a90d9 (first entry in fact block colors.bg\_by\_area)"} / {\rmfamily "text reads 'Live rankings'"}

\item
\ensuremath{\times} {\rmfamily "Implemented correctly"} / {\rmfamily "Meets the requirement"} (these are conclusions, not measurements).

\item
If the value cannot be measured from any supplied material \ensuremath{\rightarrow} write {\rmfamily "cannot confirm"} in observed.

\end{itemize}

\item[3.]
\textbf{deviation (difference)}: The difference between expected and observed. \textbf{This field must not be omitted.}
\begin{itemize}[leftmargin=1.2em,topsep=1pt,itemsep=1pt,parsep=0pt]
\item
Write {\rmfamily "none"} only when there is no difference at all. Writing {\rmfamily "none"} commits you to proving character-by-character/value-by-value equality.

\item
Otherwise, specify the difference: {\rmfamily "count differs by 2 (3/5)"} / {\rmfamily "similar hue but brightness differs by about 40\%"} / {\rmfamily "text has been reworded"}.

\end{itemize}

\end{enumerate}
\par
{\color{black!45}\texttt{[...]}}

\par\vspace{1pt}\textbf{Derive the score from deviation (do not form an impression and then justify it)}\par
\par
Complete expected / observed / deviation \textbf{first}, then use deviation to select a score from the table below. Do not reverse this order.

\par
Use a \textbf{four-level 0-1-2-3 scale} (poor / fair / good / excellent). A score of 3 is extremely difficult to earn: it requires perfect, indisputable evidence.

\par\begingroup\setlength{\tabcolsep}{4pt}
\renewcommand{\arraystretch}{1.12}
\begin{tabularx}{\linewidth}{@{}p{.13\linewidth}X@{}}
\toprule
\textbf{Level / Score} & \textbf{Meaning and corresponding deviation} \\
\midrule
\textbf{Poor}\newline \textbf{0} & Entirely missing, entirely inconsistent, clearly deviating (wrong count, reworded/summarized text, wrong position/hierarchy, clearly mismatched attributes)\newline\textit{Corresponding deviation:} One or more clear deviations, or complete absence \\
\addlinespace[3pt]
\textbf{Fair}\newline \textbf{1} & Correct direction with acceptable deviations (roughly correct but imprecise details, right color family but different values, slightly too few items, minor layout differences)\newline\textit{Corresponding deviation:} Quantifiable deviations, but recognizably the same overall design \\
\addlinespace[3pt]
\textbf{Good}\newline \textbf{2} & Essentially correct, with only tiny imperfections (color difference \textless{}5\%, spacing difference 1-2 px, almost indistinguishable)\newline\textit{Corresponding deviation:} Tiny or no deviation, but not verifiable value-by-value perfection \\
\addlinespace[3pt]
\textbf{Excellent}\newline \textbf{3} & \textbf{Perfect match}: expected and observed are equal character by character/value by value, with zero deviation and sufficient evidence\newline\textit{Corresponding deviation:} deviation is genuinely {\rmfamily "none"}, and both values can be checked character by character \\
\bottomrule
\end{tabularx}\endgroup\par
\par
{\color{black!45}\texttt{[...]}}

\end{tcolorbox}

\caption{Perceptual-judging prompt excerpts. The judge must record expected and observed values and their deviation before applying the four-level rubric.}
\label{fig:implementation-prompt-judge}
\end{figure}

\end{document}